\documentclass[acmconf,screen,nonacm,review=false]{acmart}

\usepackage{breakurl} 

\AtBeginDocument{%
  }

\setcopyright{acmcopyright}
\copyrightyear{2018}
\acmYear{2018}
\acmDOI{XXXXXXX.XXXXXXX}

\acmConference[Conference acronym 'XX]{Make sure to enter the correct
  conference title from your rights confirmation emai}{June 03--05,
  2018}{Woodstock, NY}
\acmPrice{15.00}
\acmISBN{978-1-4503-XXXX-X/18/06}

\begin{document}

\title{Five policy uses of algorithmic transparency and explainability}

\author{Matthew R.\ O'Shaughnessy}
\email{matthewosh@gmail.com}
\orcid{0000-0002-2951-8917}
\affiliation{%
  \institution{Carnegie Endowment for International Peace}
  \streetaddress{1779 Massachusetts Ave NW}
  \city{Washington}
  \state{DC}
  \country{USA}
  \postcode{20036}
}

\renewcommand{\shortauthors}{O'Shaughnessy}

\begin{abstract}
\begin{Large}
September 14, 2023
\end{Large}\\[0.25cm]
The notion that algorithmic systems should be ``transparent'' and ``explainable'' is common in the many statements of consensus principles developed by governments, companies, and advocacy organizations. But what exactly do policy and legal actors want from these technical concepts, and how do their desiderata compare with the explainability techniques developed in the machine learning literature? In hopes of better connecting the policy and technical communities, we provide case studies illustrating five ways in which algorithmic transparency and explainability have been used in policy settings: specific requirements for explanations; in nonbinding guidelines for internal governance of algorithms; in regulations applicable to highly regulated settings; in guidelines meant to increase the utility of legal liability for algorithms; and broad requirements for model and data transparency. The case studies span a spectrum from precise requirements for specific types of explanations to nonspecific requirements focused on broader notions of transparency, illustrating the diverse needs, constraints, and capacities of various policy actors and contexts. Drawing on these case studies, we discuss promising ways in which transparency and explanation could be used in policy, as well as common factors limiting policymakers' use of algorithmic explainability. We conclude with recommendations for researchers and policymakers.
\end{abstract}

\maketitle

\section{Introduction}
\label{sec:introduction}

That opaque algorithmic systems should be transparent and explainable is a common refrain among policymakers. A 2019 survey found that 73 of 84 prominent AI strategy documents referenced transparency or explainability \cite{jobin2019global}. Influential intergovernmental bodies such as United Nations agencies and the Organization for Economic Cooperation and Development (OECD) have put forth transparency and explainability as key mechanisms for ensuring that algorithmic systems produce beneficial outcomes and uphold ``democratic values'' \cite{unesco2021recommendation,oecd2020recommendation}.

Algorithmic transparency and explainability can serve many purposes, but some of the most important are legal in nature: allowing lawmakers to understand and craft effective rules for algorithmic systems, enabling a broader set of stakeholders to be aware of (and obtain redress from) algorithmic harms, and assisting regulators in exercising meaningful oversight over the use of algorithms \cite{jobin2019global,metcalf2021algorithmic}. To serve these objectives, transparency measures and explanation techniques must be developed with an understanding of the specific goals, constraints, and incentives of policymakers.

This paper aims to help bridge the gap between policymakers and the explanation research community, helping researchers to better understand and respond to the needs of policymakers. To this end, it provides case studies illustrating five uses for algorithmic transparency and explanation in policy settings. These case studies (Table \ref{tab:policy-uses}) were selected to span four axes: the spectrum from explanation to transparency (including both requirements for specific explanation techniques, like those developed by the machine learning research community, and broader forms of transparency requirements); different jurisdictions (including U.S.\ federal regulators, U.S.\ states, and the EU); policy actors with differing technical and financial capacities; and a diverse array of policy approaches (including prescriptive technical rules, process-oriented rules, nonbinding guidelines, and modifications to legal procedures).

Building on these case studies, this paper argues that explanation techniques developed by the research community can be too complex, too uncertain, or too restricted to satisfy the constraints that policymakers and the law operate under in practice. As a result, explanation is often limited in its ability to enable meaningful public policy solutions to algorithmic harms. 

We begin in Section \ref{sec:background} by contextualizing how transparency and explainability serve broader goals of algorithmic accountability, and reviewing related literature on how various stakeholders make use of explanation tools. Section \ref{sec:policy-uses} contains the main contribution of the paper: five case studies and discussion illustrating specific ways in which algorithmic transparency and explainability have been used in policy (Table \ref{tab:policy-uses}). To help researchers understand the broader context underlying each use case, we focus not only on technical details of how transparency and explainability are operationalized, but the broader policy and political context that led to the policies' development. In addition, each case study is accompanied by a discussion highlighting takeaways and other policy examples. Section \ref{sec:conclusion-recommendations} summarizes challenges surfaced in the case studies and offers recommendations for researchers and policymakers to bridge the gaps between these communities.

\renewcommand{\arraystretch}{1.2} 
\begin{table*}
	\caption{Case studies of transparency and explanation in policy}
	\label{tab:policy-uses}
	\small
	\begin{tabular}{p{0.30\textwidth}p{0.65\textwidth}}
		\toprule
		Case study & Role of transparency and explanation \\
		\midrule
		U.S.\ Equal Credit Opportunity Act (ECOA) (Sec \ref{sec:policy-uses/specific-explanations}) & Satisfy regulatory requirements for consumer-facing explanations with specific traits \\
		U.S.\ Federal Reserve Supervisory Regulation 11-7 (Sec \ref{sec:policy-uses/self-regulation}) & Enable broader and more effective challenge of algorithms' design in nonbinding guidelines for internal corporate management of algorithmic risk \\
		Colorado SB21-169 (unfair discrimination in insurance) (Sec \ref{sec:policy-uses/regulatory-approval}) & Enable compliance with governance rules applied to algorithms used in a highly regulated setting \\
		EU AI Liability Directive (Sec \ref{sec:policy-uses/liability}) & Legal guidelines intended to mitigate challenges to legal liability posed by algorithms' opacity \\
		Idaho H118 (transparency in pretrial risk assessment) (Sec \ref{sec:policy-uses/transparency}) & Broad legal requirements for model and data transparency intended to promote accountability for algorithms used in a civil-rights-critical setting \\
		\bottomrule
	\end{tabular}
\end{table*}

\section{Background: Uses and goals of transparency and explanation}
\label{sec:background}

Explanation methods are often designed by and for machine learning engineers, who use these tools to guide efforts to detect and correct problems \cite{bhatt2020machine,liao2020questioning,krishna2022disagreement} such as in accuracy, dataset composition, robustness, and fairness \cite{doshi-velez2017rigorous}. However, literature has found dissensus and confusion among data scientists who use explainability tools \cite{bhatt2020machine,kaur2020interpreting}. Data scientists typically use multiple explanation methods, often looking at sets of top-contributing features \cite{krishna2022disagreement}. When methods disagree, practitioners lack reliable heuristics for choosing between them \cite{kaur2020interpreting,bhatt2020machine,krishna2022disagreement}. They typically favor methods perceived as more consistent, theoretically-grounded, recent, or as better-aligned with the practitioner's intuition \cite{krishna2022disagreement}. While metrics have been proposed to quantitatively evaluate explanations \cite{phillips2021four}, these metrics quantify different conceptions of explainability and thus are themselves prone to disagreement \cite{krishna2022disagreement}.

However, when transparency and explainability are primarily used by developers its ability to produce accountability can be limited. Metcalf et al.\ describe three components of algorithmic accountability: 1) actors who provide information on how a system is constructed and the impacts it might have; 2) fora where this information is evaluated; and 3) changes and consequences resulting from the evaluation of this information \cite{metcalf2021algorithmic}. When used exclusively by developers, a specific type of accountability process is created in which a developer (and its engineers) act simultaneously as 1) the actor producing information about algorithmic systems; 2) the forum evaluating this information and judging its merits; and 3) the decision-maker setting and justifying the resulting course of action. By allowing developers to create and enforce their own policies, this setup can lead to several failures of meaningful accountability. As described by Metcalf et al.\ \cite{metcalf2021algorithmic}, it lacks a external forum to force changes; suffers from conflicts of interest; is obscured from public scrutiny; reduces creation of broadly-recognized norms; and reduces input from a broad range of expertise and impacted users \cite{moss2021assembling,porter2022distinguishing}.

Transparency and explainability can also provide utility to stakeholders beyond developers. It can promote an understanding of algorithmic systems that can enable public participation and broader discussion about when and how algorithmic systems should be used \cite{jobin2019global,oecd2020recommendation}. It can help end-users understand when to trust algorithmic systems' outputs, when to question them, and when to intervene on them \cite{bhatt2020machine,liao2020questioning,hong2020human,amarasinghe2021explainable}. Explainability and interpretability has been suggested as a necessary precursor to user trust and adoption \cite{hall2019introduction,hong2020human,liao2020questioning,zerilli2022how,amarasinghe2021explainable}. Literature has described the utility of explanation in contexts from medical \cite{elshawi2019interpretability} to financial \cite{zhang2022explainable,mcelfresh2022machine}.

But existing explanation techniques also have many important limitations. By describing only a small piece of the operation of complex algorithmic systems, they can mislead their users \cite{neely2021order,krishna2022disagreement}. Exacerbating this problem, explanations themselves are often sensitive to perturbation, inaccurate \cite{kindermans2019reliability}, or provide the false impression that a system is robust or free of bias \cite{aivodji2019fairwashing,lakkaraju2020how}. Commentators argue that explanations are rarely tailored to the audience that they are consumed by \cite{tomsett2018interpretable,arya2019one,bhatt2020explainable,amarasinghe2021explainable,hoffman2021stakeholder}, and can shift the burden for obtaining recourse to individuals \cite{whittaker2018ai}. For developers, explanations can come at the cost of performance metrics such as accuracy \cite{doshi-velez2017accountability}, expose proprietary details, or pose privacy risks \cite{phillips2021four}.

\section{Five uses of explainability in policy}
\label{sec:policy-uses}

This section contains in-depth case studies illustrating the political and policy considerations underlying five recent regulatory and legislative proposals involving algorithmic transparency and explainability (Table \ref{tab:policy-uses}). The case studies were developed through literature review, records of policy processes, and interviews with key stakeholders.

\subsection{Use \#1: Specific explanation requirements}
\label{sec:policy-uses/specific-explanations}

Policymakers often see the provision of explanations as a tool to encourage decision-making to be more transparent, fair, and accountable \cite{doshi-velez2017accountability,deeks2019judicial,phillips2021four}. The U.S.\ Equal Credit Opportunity Act (ECOA) is one such regulation requiring important decisions to be accompanied by explanations with certain characteristics \cite{brotcke2022time}.

\subsubsection{Case study: U.S.\ Equal Credit Opportunity Act}
\label{sec:policy-uses/specific-explanations/case-study}

Credit is a ``foundational drive[r] of opportunity and equality'' \cite{akinwumi2021ai}, making the decision-making processes that determine when and how people are eligible for credit an important social problem \cite{rice2013discriminatory,barocas2016big}. In response to widespread credit discrimination, in 1974 the U.S.\ Congress passed ECOA, which made it unlawful for creditors to discriminate on protected bases such as race, sex, and marital status \cite{unitedstates1974equal,1976ecoa}. A critical provision of ECOA and Regulation B, the detailed technical rules developed to implement the law \cite{consumerfinancialprotectionbureau201812}, is that creditors taking ``adverse actions'' such as denying credit must provide consumers with a ``statement of specific reasons for the action'' \cite[\textsection9(a)(2)]{consumerfinancialprotectionbureau201812}.

The specific requirements described in Regulation B illustrate the level of detail that these rules can entail. Explanations must be ``specific'' and ``indicate the principal reason(s) for the adverse action'' rather than describing the creditors' general policies \cite[\textsection9(b)(2)]{consumerfinancialprotectionbureau201812}. These reasons must ``accurately describe the factors actually considered or scored'' and ``no factor that was a principal reason for adverse action may be excluded from disclosure,'' but creditors need not describe exactly how factors affected the application \cite[Comment for \textsection9(b)(2)]{consumerfinancialprotectionbureau201812}. A set number of these reasons is not required, though the regulation suggests that ``disclosure of more than four reasons is not likely to be helpful to the applicant''\cite[Comment for \textsection9(b)(2)-1]{consumerfinancialprotectionbureau201812}.\footnote{Highlighting the high degree to which specific explanation requirements in Regulation B are parsed, the Fair Credit Reporting Act (FCRA) requires subtly different types of explanations that are often discussed in conjunction with Regulation B's. FCRA, for instance, explicitly requires that explanations contain no more than four ``factors,'' and some analysts have suggested that ECOA's required ``reasons'' and FCRA's required ``factors'' may subtly differ. See, e.g., \cite{ammermann2013adverse,pace2022dark}.} Official interpretive notes state that ``[v]arious methods will meet the requirements'' \cite[Comment for \textsection9(b)(2)-5]{consumerfinancialprotectionbureau201812} and provide quasi-algorithmic examples illustrating satisfactory explanations. For instance, one sample explanation method is to ``identify the factors for which the applicant's score fell furthest below the average score for each of those factors achieved by all applicants'' \cite[Comment for \textsection9(b)(2)-5]{consumerfinancialprotectionbureau201812}. Potential factors suggested include ``length of employment,'' ``limited credit experience,'' and ``number of recent inquiries on credit bureau report'' \cite[Appendix C]{consumerfinancialprotectionbureau201812}.

Legislators, regulators, and courts have described two major goals for ECOA's explanation requirements. The first is to educate consumers, allowing them to take steps to improve their credit or fix errors \cite{94thcongress1976senate,u.s.federaltradecommission2013ftc}. The second is to discourage discrimination, as articulated by the U.S.\ Senate in 1976: ``if creditors know they must explain their decisions ... they [will] effectively be discouraged from discriminatory practices'' \cite{94thcongress1976senate,collazo2021advanced}. There is some debate about whether ECOA's explanation requirements are effective in this goal. An officer of one creditor has argued that they provide accountability, forcing algorithm developers to ask themselves when selecting variables, ``[a]re you comfortable telling the world that this is what you're relying on?'' \cite{upbin2020how}. Other analysts, however, have argued that adverse action notices are rarely actionable or consumer-friendly, and thus provide limited practical value to consumers \cite{akinwumi2021ai}.

The Consumer Financial Protection Bureau (CFPB), the regulator that enforces Regulation B, has shifted its stance on how its explanation requirements apply to algorithmic decision systems as political leadership has changed. During the Trump Administration, the CFPB portrayed the increasing use of algorithms in credit as an innovation that could increase equality and reduce costs. Noting that ``industry uncertainty about how AI fits into the existing regulatory framework may be slowing its adoption, especially for credit underwriting'' \cite{ficklin2020innovation}, the CFPB provided protection against enforcement to a company experimenting with increased use of external data for making credit decisions \cite{ficklin2019update}, arguing that the ``built-in flexibility'' of Regulation B ``can be compatible with AI algorithms'' \cite{ficklin2020innovation}.

This accommodating tune changed during the Biden administration, when the CFPB placed a greater emphasis on algorithmic decision-making's potential to produce discriminatory outcomes. The CFPB's new director warned creditors of a new emphasis on how ``so-called neutral algorithms ... may reinforce biases that have long existed'' \cite{2021remarks}, and the Trump-era ``Office of Innovation'' that had issued more flexible algorithmic regulatory approaches was replaced \cite{2022cfpb}. Most notably for explanation requirements, in May 2022 the CFPB published a policy statement emphasizing that creditors ``cannot justify noncompliance with ECOA and Regulation B's requirements based on the mere fact that the technology it employs to evaluate applications is too complicated or opaque to understand'' \cite{consumerfinancialprotectionbureau2022consumer}. The statement, called CFPB Circular 2022-03, explicitly described a broad class of post-hoc machine learning explanation methods as potentially noncompliant, stating that explanations based on ``approximate models'' may not satisfy Regulation B's accuracy requirements for explanations \cite[Footnote 1]{consumerfinancialprotectionbureau2022consumer}.

Even before this stricter interpretation, though, the explanation requirements in ECOA and Regulation B have strongly shaped the use of algorithms in credit underwriting. While analysts have argued that many post-hoc explanation techniques can comply with Regulation B \cite{hall2019introduction,pace2022using,krivorotov2022explaining}, these rules add to creditors' caution when using algorithmic tools, potentially leading them to favor the use of more interpretable models for key decision-making tasks and creating a market for algorithmic tools specifically designed to satisfy Regulation B \cite{hall2021united}.

\subsubsection{Discussion and other policy examples}
\label{sec:policy-uses/specific-explanations/discussion}

Despite being written long before the use of algorithms became widespread in credit, the explanation requirements in Regulation B now apply to decisions that are frequently made by algorithms. This illustrates a central tenet in technology policy: technology often outpaces the development of policy responses \cite{marchant2011addressing}. This ``pacing problem'' can be particularly pernicious in specific requirements for machine learning explanations, where both algorithms and state-of-the-art explanation methods evolve faster than policy ordinarily can. The more flexible policy uses of explanation described in the remaining case studies offer the advantage of being adapt more quickly as technology evolves.

Another common specific policy requirement for explanations is that they be understandable to the public, a call frequently cited as supporting public participation and democracy \cite{jobin2019global,oecd2020recommendation,bradley2021national}. For instance, recommendations for the AI industry developed by the city of Shenzhen call for the provision of explanations that are ``conducive to public understanding'' \cite[Article~71]{shenzhenmunicipalpeoplescongress2022shenzhen}. Illinois's 2019 ``Artificial Intelligence Video Interview Act'' requires that companies using machine learning tools to evaluate video interviews provide ``information before the interview explaining how the artificial intelligence works and what general types of characteristics it uses to evaluate applicants'' \cite{illinoisgeneralassembly2019artificial,neace2022aivia}.

\subsection{Use \#2: Nonbinding guidelines for internal algorithmic governance}
\label{sec:policy-uses/self-regulation}

In ``soft law'' governance approaches, regulators create flexible guidelines for how algorithms should be used rather than defining precise binding rules \cite{gutierrez2021global}. Guidance on mitigating algorithmic risk from U.S.\ financial regulators provides one example of how transparency and explanation can be used in this form of policy.

\subsubsection{Case study: U.S.\ Federal Reserve SR 11-7 --- ``Guidance on model risk management''}
\label{sec:policy-uses/self-regulation/case-study}

Financial institutions are prolific users of algorithmic systems. Particularly in the wake of the 2008 financial crash, their industry is closely overseen by regulators monitoring for systemic risks \cite{crespo2017evolution}. U.S.\ financial regulators' Supervisory Regulation (SR) 11-7, ``Guidance on Model Risk Management,'' provides nonbinding guidelines on how financial institutions such as banks should safeguard against risks introduced by incorrect or misused models \cite{federalreserve2011guidance}.

SR 11-7 is described as ``guidance'' rather than strict regulation,\footnote{SR 11-7's use of flexible rather than precise requirements is distinct from the more controversial question of whether the guidance as a whole can be formally described as a ``rule,'' a technical distinction that determines whether it is subject to the Congressional Review Act \cite{armstrong2019board}.} and indeed it is intended to be applied flexibly based on organizations' ``size, nature, and complexity, as well as the extent and sophistication of [their] use of models'' \cite{federalreserve2011guidance}. While it contains some level of prescriptive detail \cite{kiritz2018supervisory}, SR 11-7 does not attempt to define precise requirements that apply across all contexts and businesses \cite{u.s.officeofthecomptrollerofthecurrency2021comptroller}. As a result, it can be significantly more technically nuanced than the rigid requirements outlined in more formal regulation and legislation.

Over 21 pages of detailed guidance, the document focuses on three phases of oversight \cite{federalreserve2011guidance}. The first describes the development, implementation, and use of models themselves. Noting that the design process is inherently subjective, here the guidance emphasizes accuracy and robustness testing as well as detailed expectations for transparency in procedural steps, such as the documentation of design choices, data sources, and assumptions. The second phase covers model validation. This describes analyzing, documenting, and monitoring model conceptual soundness, limitations, assumptions, data relevance, and robustness. Reflecting the guidance's flexible nature, validation is recommended to be outcomes-oriented rather than a ``straightforward, mechanical process that always produces unambiguous results'' \cite[pg.~15]{federalreserve2011guidance}. The final phase of oversight described by SR 11-7 describes governance processes \cite{federalreserve2011guidance}, such as the clear delineation of responsibilities between business units and personnel, the role of boards and risk management offices, and audit policies \cite{deloittecenterforregulatorystrategy2018model}. Meaningful documentation is also described as a key component of oversight. Attention is given to incentivizing developers to create robust documentation, which might include explanation.

SR 11-7 emphasizes sociotechnical and organizational factors inherent to robust model design and validation. A ``guiding principle'' throughout the document is the development of organization-wide capability for ``effective challenge'' of models, described as requiring not only technical capacity but organizational structures that allows criticism to have meaningful influence. To this end, the guidance recommends steps to incentivize the challenge of models at every stage of development and use, and to ensure it comes from independent teams with both diverse experience and the influence to create change. As with technical aspects of the guidance, governance recommendations are described not as ends to themselves, but as strategies for encouraging more effective risk management.

Transparency and explainability are highlighted in many parts of SR 11-7, and explanation is especially well-suited to the type of flexible expectations laid out by SR 11-7. In contrast to regulatory settings where policymakers define precise and rigid rules that models must follow --- where unreliability and lack of robustness can limit explanation's usefulness for demonstrating compliance --- SR 11-7 embraces the more exploratory evidence that a battery of explanation techniques could provide. A recommendation on the use of statistical testing, for example, applies equally well to explanation techniques: ``in many cases statistical tests cannot unambiguously reject false hypotheses or accept true ones based on sample information. Different tests have different strengths and weaknesses under different conditions. Any single test is rarely sufficient, so banks should apply a variety of tests to develop a sound model'' \cite[pg.~6]{federalreserve2011guidance}. The guiding principle of ``effective challenge'' also provides an ideal framework for the use of explanation to test aspects such as robustness and the potential for data to serve as a proxy for protected variables. Finally, model explanations may play a central role in the extensive documentation recommended by SR 11-7, where several pieces of individually-inconclusive evidence could together justify certain modeling decisions.

\subsubsection{Discussion and other policy examples}
\label{sec:policy-uses/self-regulation/discussion}

SR 11-7's focus on governance strategies rather than precise technical requirements bears similarities to the Colorado Division of Insurance's proposed regulation (Section \ref{sec:policy-uses/regulatory-approval/case-study}). While differences exist in regulator capacity and sector-specific traditions, both case studies illustrate how transparency and explanation can fit neatly into flexible guidance on process and governance, not just into precise technical requirements. From a regulator's perspective, regulating process and governance is helpful because it does not require developing complete understanding of how complex models work \cite{deloittecenterforregulatorystrategy2018model}. It also avoids the drawbacks of placing burdensome technical constraints (such as those illustrated in Section \ref{sec:policy-uses/specific-explanations}) on how algorithms are used.

This focus on governance is shared by many other policy efforts, and explainability can fit naturally into these strategies. The AI Risk Management Framework developed by the U.S.\ government, which suggests best practices for a broad range of developers, contains a section on explainability \cite[Sec.~3.5]{u.s.nationalinstituteofstandardsandtechnology2023artificial}, and its final version will contain a detailed playbook on how models should be ``explained, validated, and documented'' \cite[``MEASURE''~2.9]{u.s.nationalinstituteofstandardsandtechnology2023artificial}. The European Commission's draft request for standards to implement the EU AI Act focuses on governance aspects such as ``risk management,'' ``data quality and governance,'' and ``technical documentation'' in addition to more technical aspects \cite{europeancommission2022draft}.

Such a focus might be most effective when the goals of regulators and the broader industry are aligned and regulation can focus on ensuring that best practices maintained by leading companies are more universally performed. Some argue that this is the case in finance, where neither regulators nor major banks directly benefit from systemic risk or misused models \cite{crespo2017evolution,deloittecenterforregulatorystrategy2018model,andrews2008greenspan}. In sectors and contexts where regulators and industry tend to be more opposed, however, flexible regulation or a governance-oriented strategy to algorithmic regulation might be less likely to succeed.

\subsection{Use \#3: Governance rules in highly regulated settings}
\label{sec:policy-uses/regulatory-approval}

In so-called ``highly regulated industries'' such as healthcare and insurance, regulators have the authority to place strict rules on the use of algorithmic tools. In many cases, this can influence when and how explanation is used.

\subsubsection{Case study: Colorado SB21-169 (2021) --- ``protecting consumers from unfair discrimination practices''}
\label{sec:policy-uses/regulatory-approval/case-study}

Insurance is one such highly regulated industry in the U.S., and as a consequence insurers view algorithmic and machine learning tools with both optimism and trepidation. On the one hand, insurers view data as offering a valuable competitive advantage: major insurers have large data science teams \cite{klein2019company}, and metrics based on credit scores, for instance, are seen by insurers as highly predictive of losses \cite{frees2021discriminating}. On the other hand, insurers' use of data and predictive models are often shaped by regulatory and reputational considerations. In some states, regulators must approve insurers' methods for determining rates before plans can be written. This approval process often prohibits the use of certain variables or proxies \cite{frees2021discriminating}, though exact details vary widely by state and insurance type.

Insurers' use of data and algorithms is also shaped by reputational considerations. Insurers worry that asking for sensitive information such as social security numbers or protected characteristics like race will scare customers or weaken their ability to make ``fairness through unawareness''-type arguments if accused of bias. As a result, insurers are often incentivized to use simpler and more interpretable data and models in practices that are subject to greater scrutiny, even as they use more complex models for practices that are less regulated, such as advertising or assigning claims adjusters to cases.

Against this backdrop, in 2021 the Colorado State Legislature passed SB21-169, which targets insurers' use of predictive models and ``external consumer data and information sources'' (ECDIS) in ways that result in ``unfair discrimination'' against protected classes \cite{buckner2021concerning}. The bill, whose provisions were generally quite broad \cite{black2021insurers}, instructs Colorado's Division of Insurance to begin a rulemaking process to develop more prescriptive rules governing the use of ECDIS in predictive algorithms. The law also mandates that the Division use an open stakeholder consultation process, providing an opportunity to better understand policymakers' motivations and constraints.

In initial stakeholder consultation meetings, the Division described the law's motivation by sharing examples of racial and gender gaps in auto \cite{angwin2017minority}, life \cite{medine2020there}, and homeowners insurance \cite{flitter2020black}, and by pointing to work demonstrating similar disparities in facial recognition technology \cite{buolamwini2018gender} \cite{coloradodivisionofinsurance2022slide}. The Division emphasized its intention to encourage the ``responsible'' use of big data rather than prevent its use, as well as its hope that by developing rules slowly and carefully its regulatory measures would eventually serve as a model for other states \cite{stateofcolorado2022division}. The Division also emphasized its desire to develop a regulatory structure that would be efficient for both insurers and regulators \cite{coloradodivisionofinsurance2022slide}.

Indeed, the Division's limited capacity and technical expertise --- as in other states, the regulator is responsible for overseeing hundreds of insurers with a relatively limited staff and budget --- has been a central concern. The law requires that insurers ``provide [the Division] an explanation of the manner in which [they use] ECDIS'' \cite{buckner2021concerning}, though this requirement refers to a general description of algorithms' purpose and use rather than specific technical explanations. In a stakeholder meeting the Division stated that ``outcomes derived through the use of ECDIS, algorithms, and predictive models must have a rational explanation and be explainable internally as well as to regulators and the ... public,'' noting that transparency and explainability (broadly construed) are important components of maintaining public trust in the insurance industry \cite{coloradodivisionofinsurance2022recording}. But the Division, like most other state regulators, is unlikely to have the resources to evaluate explanations or technical details made available by transparency measures in a way that would enable a judgment on whether a particular algorithmic practice would result in ``unfair discrimination.''

Despite this, explanations as conceived by the machine learning research community could be used internally to help companies comply with the Division's articulated focus on testing and outcomes. While final rules are still under development, initial discussion has focused on outcomes-based testing with a variant of the BISG algorithm \cite{coloradodivisionofinsurance2022recordinga,coloradodivisionofinsurance2022recordingb,stateofcolorado2022division} commonly used in other regulatory settings, which allows for testing of racial disparities even when race data is not collected by inferring it with names and addresses \cite{randcorporation2021bayesian,u.s.consumerfinancialprotectionbureau2014using}. When these types of tests are encouraged or required, explanation can serve as an important internal tool to aid developers in complying with outcomes-based rules.

Transparency and explanation may also be important elements of compliance with potential governance requirements proposed by the Division \cite{coloradodivisionofinsurance2022recordingc}. These proposed rules would include aspects such as the creation of formalized multi-disciplinary oversight teams and more formalized structures for accountability and oversight. They would also include requirements for documenting the justification for when and how algorithms are used, trained, monitored, and updated \cite{coloradodivisionofinsurance2022recordingc}. If enacted, these documentation and reporting requirements could force algorithm developers' use of explanation during the model development process to become more formalized.

\subsubsection{Discussion and other policy examples}
\label{sec:policy-uses/regulatory-approval/discussion}

The Colorado law reflects a growing awareness that ``fairness through unawareness'' is insufficient in industries such as insurance \cite{krafcheck2021insurance}, where decisions can have powerful impacts on lives \cite{barocas2016big}. Colorado's Insurance Commissioner has stated that he hopes that Colorado's effort will serve as a model for other states \cite{coloradodivisionofinsurance2022recordinga}, and indeed similar bills have already been introduced in other legislatures \cite{oklahomastatelegislature2022bill,rhodeislandlegislature2022legislative}. While regulatory philosophies differ across states, if the law is seen as successful  mechanisms like the National Association of Insurance Commissioners' database of model laws\footnote{https://content.naic.org/model-laws} could encourage development of similar laws and regulations.

The Colorado law also demonstrates the substantial role that reputational concerns play in governing industries adopting algorithmic tools. For example, reputational concerns are likely significant factors in insurers' hesitation to collect personal information that might help them test for the existence of bias in their tools. Companies are likely especially cautious around topics like fairness, where different stakeholders view the problem very differently and conflicting definitions abound. Companies that see themselves as unable to please everyone and protect themselves from all accusations of using discriminatory algorithms may seek to protect themselves by withholding as much information as possible, potentially jeopardizing consumers' and advocacy organizations' ability to understand how companies use algorithmic tools.

Transparency often comes into tension with consumer privacy and company intellectual property \cite{bogen2020awareness,phillips2021four}. In Colorado SB21-169, information provided by insurers to the Commissioner's office is treated as proprietary, allowing the Commissioner to use it for regulatory purposes but shielding it from disclosure to the public \cite[Section 2(3)(d)]{buckner2021concerning}. This benefits insurers, who jealously guard details of how they use predictive algorithms --- but also protects customers from privacy violations. However, these data protections mean that, unlike regulation that provides broad transparency (e.g., see Section \ref{sec:policy-uses/transparency}), individuals and civil society have a limited ability to affect accountability from insurers on their own; instead, they must rely on the regulator to detect problematic uses of ECDIS from insurers' filings. Other highly-regulated industries take alternate approaches to this trade-off. For example, federal regulators require some lending companies to collect demographic data from mortgage applicants to enable monitoring for discrimination \cite{williams2008fair}.

The exact form and strength of regulatory oversight differs dramatically across industries, products, and jurisdictions. Health insurance is typically subject to stricter rules than life insurance, for example, and regulatory philosophies differ across U.S.\ states and countries. Another setting in which explanation plays a role in a highly regulated environment is discussed in Section \ref{sec:policy-uses/self-regulation}.

\subsection{Use \#4: Enable or interface with legal liability}
\label{sec:policy-uses/liability}

\subsubsection{Liability as a form of algorithmic governance}
\label{sec:policy-uses/liability/background}

Liability has an important influence on how technology is developed and used. For companies and developers, liability risk looms large when deciding whether to bring a product to market, thus discouraging risky behavior. For policymakers, liability can be an attractive regulatory tool because decisions ultimately made by courts and juries can be much more context-aware than legislation \cite{cuellar2019common}. Liability can be effective in producing change: in the U.S., vaccines have been removed from the market because of liability risk \cite{aronowitz2012rise}, and commentators have argued that liability risks helped guide the development of new nanotechnology products by allowing reinsurance companies to serve as ``quasi-regulators'' \cite{tournas2021ai}.\footnote{But see also \cite{abraham2022limits} on the limitations of liability insurance as a tool for shaping behavior.}

The law establishes what actions defendants can be liable for, what plaintiffs or complainants must prove to obtain redress, and what information defendants must release to plaintiffs. Details differ across legal systems and are subject to court interpretation, but typically involve factors such the strength and directness of the causal link between a defendant's action and the harm, the foreseeability and preventability of a harm, and the defendant's intent \cite{cuellar2019common,cardi2011hidden,bathaee2018artificial}.

Policymakers adjust these factors to control how hard it is for litigation to succeed, thus allowing them to dictate how strongly to discourage risky behavior in various settings.\footnote{This tool, of course, is delicate. For example, Bathaee argues that to be effective at moderating risky behavior, stronger types of liability should be used in areas where actors have some degree of predictability of potential risks (e.g., for insurance companies to establish rates), and that liability can impose major barriers to market entry that can unfairly advantage larger companies \cite{bathaee2018artificial}.} For example, the law discourages ``ultrahazardous activities'' (such as the unnecessary storage of explosives near pedestrians) by subjecting them to ``strict liability,'' in which it is not necessary to prove that the defendant was at fault or intended for a harm to occur \cite{wex2022ultrahazardous}. This creates a strong incentive to abstain from or reduce the risks of activities dubbed ultrahazardous. The law can also make litigation \emph{less} likely to succeed by increasing the burden of proof \cite{bathaee2018artificial}. For example, parts of U.S.\ antidiscrimination law require that plaintiffs not only show that disparate impact occurred, but also that the plaintiff \emph{intended} for the discrimination to occur --- thus making it more difficult for discrimination cases to succeed \cite{huq2018what}. By contrast, types of liability in which plaintiffs need only show that the defendant was negligent (for example, by not following accepted best practices) make it easier for plaintiffs' cases to succeed \cite{marchant2019ai}.

However, it can be more difficult for victims to obtain redress through liability for algorithmic harms than for traditional products. Decisions based on complex rules learned from data coupled with a lack of transparency can make it difficult for plaintiffs to demonstrate a causal link between an algorithmic system and a harm --- or to realize that a harm occurred in the first place \cite{ecexpertgroup2019liability,wendehorst2022ai}. The separation or removal of humans from decision making can make it difficult to prove fault or intent; training algorithmic systems with biased datasets, for example, has not typically seen by U.S.\ courts as demonstrating intent to harm \cite{bathaee2018artificial}. The complexity of many algorithmic systems means that interrogating them in court requires expensive expert analysis, potentially making it prohibitively costly to litigate all but the most well-funded or obvious cases \cite{bathaee2018artificial,ecexpertgroup2019liability}.

Algorithmic transparency and explanation can help address these concerns \cite{deeks2019judicial}, but algorithms and explanations are often nuanced, technically complex, and require expertise to analyze \cite{tomsett2018interpretable,poursabzi-sangdeh2021manipulating,krishna2022disagreement}. One way in which they could interface with liability is through expert witness analysis, in which experts would interpret aspects of models, datasets, or documented explanations and testify to a court. Such expert witness analysis could be useful for establishing a general sense of an algorithm and its development procedure, providing evidence on whether developers or operators were taking ``reasonable'' precautions or following accepted practices when designing and deploying opaque algorithmic systems, just as in a medical malpractice case an expert might testify on whether a doctor adhered to professional medical norms \cite{priceii2018medical}.

\subsubsection{Case study: EU AI Liability Directive}
\label{sec:policy-uses/liability/case-study}

In the EU, claimants normally must prove liability by demonstrating both the existence of a \emph{fault} (i.e., negligence or intent to harm) and a \emph{causal link} from that fault to a harm. The European Commission's proposed ``AI Liability Directive,'' released in September 2022, is an attempt to mitigate the unique challenges that make these two concepts difficult to prove in the presence of algorithmic systems. It does so through two principle mechanisms.

First, the proposed directive would allow greater access to the evidence about algorithmic systems a claimant would need to meet their burden of proof, allowing courts to order disclosure of information (such as documentation or logs) that are ``necessary and proportionate'' to prove ``plausible'' claims \cite[Article~3]{europeancommission2022proposal}. These disclosure requirements are meant to counteract the ``information asymmetries'' \cite{ecexpertgroup2019liability} and opacity concerns introduced by algorithmic systems.

Second, the proposed directive introduces, in certain cases, a ``rebuttable presumption'' that a demonstrated fault had a causal link to a harm \cite[Article~4]{europeancommission2022proposal}. When this ``presumption of causality'' applies, a claimant must still demonstrate the defendant's fault (for example, non-compliance with national, EU, or court disclosure requirements) --- but instead of needing to prove that the fault had a causal link to a harm, the claimant would need only demonstrate that the causal link is ``reasonably likely.'' The defendant would then bear the higher burden of \emph{disproving} the existence of this causal link. As with the disclosure requirements, this rebuttable presumption of causality is meant to be applied in a manner proportionate to the amount that the complexity and opacity of an algorithm makes proving liability more challenging.\footnote{This approach is not unique in European law. For example, in medical malpractice cases, courts can ``place the burden of producing evidence on the party who is or should be in control of the evidence'' \cite{ecexpertgroup2019liability}. In a scientifically complex case relating to vaccine damage, a court granted claimants a ``rather far-reaching presumption of causation'' \cite{ecexpertgroup2019liability}.}

These provisions could make explanations both more readily available and more likely to bear weight in court. First, it could incentivize algorithmic transparency because ``maintaining robust documentation of model testing'' will help operators of algorithmic systems avoid liability under the directive \cite{gesser2022eu}. Although the text of the proposed directive does not explicate specific factors, the expert group recommendation that led to its drafting suggests that courts weigh the ``degree of ex-post traceability and intelligibility'' of an algorithmic system, as well as the ``accessibility and comprehensibility'' of the data it used, when deciding whether to modify the burden of proving causality \cite{ecexpertgroup2019liability}. Second, by reducing the burden of proof in certain settings, the proposed directive would increase the ability of less-than-definitive evidence, such as explanations of opaque algorithms, to bear on a court's thinking.

\subsubsection{Discussion}
\label{sec:policy-uses/liability/discussion}

Liability reform illustrates how algorithmic regulation is often driven by a confluence of business, regulator, and consumer interests. Like other EU AI regulatory action, the proposed liability directive is motivated not only by protecting consumers from algorithmic harms, but also by the economic and business benefits brought by harmonizing legal standards in the 27 individual EU countries \cite{karner2021comparative,deloitte2021study,europeancommission2022proposal,europeancommissiondirectorate-generalforcommunicationsnetworkscontentandtechnology2020european}. The liability directive is also intended to increase the power of the EU's AI-related legislation \cite{europeancommission2021proposal} and the EU Charter of Fundamental Rights by allowing private right of action for violations \cite{wendehorst2022ai} \cite[pp.~9-10]{europeancommission2022proposal}.

In most cases algorithmic explanations (and thus testimony based on them) do not precisely and conclusively demonstrate causal relationships between an algorithm and a specific harm, making it unlikely that they could be dispositive in proving the kinds of causal relationships required in liability cases. Rather, explanations would typically render a strictly opaque algorithm as what Bathaee describes as ``weak black-box'': an algorithm accompanied by ``limited and imprecise'' information about its operation \cite{bathaee2018artificial}. This type of analysis also requires the availability of models, data, and explanations from the development and use of algorithmic systems, a critical aspect of the proposed EU liability directive \cite{europeancommission2022proposal}.

However, the lack of widely-accepted best practices for how explanations should be used and documented during algorithm development makes it less likely that an expert witness could convincingly point to any specific developer behavior as significantly deviating from professional norms. This might even include a developer not using explanation techniques at all during a development process, as the defendant could simply argue that the lack of consistency and reliability of explanation tools makes it unclear how their output should be interpreted (see Section \ref{sec:background}). Current business and regulatory efforts to develop these types of norms (e.g., \cite{u.s.nationalinstituteofstandardsandtechnology2023artificial,arya2019one}) could aid the utility of explanation in liability.

\subsection{Use \#5: Broad requirements for model and data transparency}
\label{sec:policy-uses/transparency}

\subsubsection{Case study: Idaho H118 (2019) --- ``to provide certain requirements for pretrial risk assessment tools''}
\label{sec:policy-uses/transparency/case-study}

Pretrial risk assessment tools are algorithmic systems used to inform judges' decisions to hold people accused of crimes in jail until trial based on a statistical estimate of the accused's risk of committing a new crime or not reappearing in court \cite{desmarais2019pretrial}. While their advocates argue that pretrial risk assessment tools allow for reduced incarceration, others have pointed to the risk that these tools' statistical nature can result in bias against historically overincarcerated groups, and that their opaque nature violates the due process rights of defendants \cite{eckhouse2019layers} (see \cite{nationalconferenceofstatelegislatures2022racial}). ProPublica's investigation of one such system, the COMPAS algorithm, helped draw policy and research attention to algorithmic fairness and bias \cite{angwin2016machine}.

Against this backdrop, in 2019 the Idaho Legislature passed House Bill 118, which imposed some of the strongest broad transparency requirements on pretrial risk assessment tools in the U.S.\ \cite{idaholegislature2019house}. The law mandates that ``[a]ll documents, data, records, and information'' used in the creation of pretrial risk assessment tools must be ``open to public inspection, auditing, and testing'' \cite{idaholegislature2019house}. The law also restricts risk assessment tool developers from using trade secret or intellectual property protections to withhold information, a provision inspired by a court case in which trade secret protections were used to justify not providing a defendant information about how their COMPAS score was computed \cite{bradley2016state,whittaker2018ai}.

Idaho H118's imposition of broad transparency requirements, rather than more narrow rules for the developers or users of pretrial risk assessment tools create explanations of their operation, was likely a product of two factors. First, the Idaho bill was developed without significant technical input \cite{idaholegislature2019proceedings,idaholegislature2019proceedingsb,idaholegislature2019proceedingsa}, so even if lawmakers had wanted to mandate explanations like those developed by the research community, they may have been unaware of this possibility. Second, the law's approach to transparency provides information that is meant to be consumed by fundamentally different audiences --- defense lawyers, research groups, and civil society watchdogs --- than policy proposals to directly help individual users to understand how algorithms impact them. As the bill's sponsor argued, ``if [pretrial risk assessment algorithms] are going to be used ... they need to be where they can be challenged and [watchdogs] hopefully expose them. If [problems such as bias] present themselves,'' outside entities should be able to ``push the issues'' \cite{lipton2019idahoa}.

The Idaho bill's use of transparency requirements was also the result of political compromise: as originally introduced the bill contained not only transparency provisions, but also a blanket prohibition on tools that were not ``free from bias'' \cite[see Original Bill Text]{idaholegislature2019house}. In an initial committee hearing, however, this bias prohibition received pushback on both technical and policy grounds \cite{idaholegislature2019proceedings}. A Computer Science professor testifying in support of the transparency aspects of the bill explained to lawmakers the difficulties of conflicting definitions of fairness and tensions with individual privacy. Another testimony questioned who would assess whether pretrial risk assessment tools were ``free from bias,'' and who would pay for these assessments. While the bill sponsor claimed that a majority of Idaho counties used pretrial risk assessment tools, it was unclear whether any of these counties were actually using the types of commercially-produced opaque algorithms that had drawn media attention, rather than simpler checklist-type assessments \cite{konig2021whenb}.

Other lawmakers expressed policy and political objections to the original bill \cite{konig2021whenb}. Some witnesses argued that nationwide debate about pretrial risk assessment tools did not apply to Idaho, and one lawmaker questioned the objectivity of ProPublica's COMPAS investigation \cite{idaholegislature2019proceedings}. Lawmakers and some witnesses expressed confusion about how the risk assessment tool used in one Idaho county, a simple questionnaire consisting of eight equally weighted and nominally race neutral questions (see \cite{rao2016transition}), could be racially biased. Others argued that lawmakers should not interfere with judges who found risk assessment scores useful. A recurring theme of the committee debate centered on the tension between individual rights and efficiency. In an op-ed, a former Idaho sheriff argued that predictive algorithms could increase efficiency, likening the practice of using factors correlated with increased risk to automotive insurers setting higher rates for teenage drivers \cite{raney2019guest}. The bill's sponsor argued strongly against this efficiency-based view, stating that ``individual rights are at the heart of our criminal justice system, not efficiency'' \cite{idaholegislature2019proceedings,chaney2019criminal}. These policy disagreements were never fully resolved.

The restriction of the bill to its transparency-focused components thus allowed policymakers to agree on a meaningful policy prescription even as they disagreed on the nature, extent, or even existence of a problem \cite{lipton2019idahoa}. The original bill containing both transparency provisions and a prohibition on ``bias'' was contentious and only narrowly voted out of committee. By contrast, once stripped to its transparency-focused components the bill passed the Idaho legislature nearly unanimously and was signed into law by the Governor \cite{idaholegislature2019house}.

While the law's attempt to ensure that developers will make publicly available the tools needed for independent researchers to generate explanations, neither the originally-introduced or amended version of the Idaho bill required that developers provide explanations of pretrial risk assessment algorithms' operation or limitations. (By contrast, in \emph{State v.\ Loomis} the Wisconsin Supreme Court ruled that cases using the COMPAS algorithm must include a ``written advisement listing [its] limitations'' \cite{bradley2016state}.)

The sponsor of the bill has indicated that an impact of the law might be to shift the development of future algorithmic tools from for-profit companies to (presumably more socially-oriented) nonprofits \cite{chaney2019criminal}. The law also allows Idaho defendants who believe that predictions about them made by pretrial risk assessment algorithms are incorrect to obtain more information about their use (see, e.g., \cite{washington2018how}). However, it is difficult to empirically assess H118's impact in the three and a half years since it was passed. The largest county in Idaho continues to use a revised version of its simple risk assessment questionnaire \cite{adacountysheriffsoffice2019ada,konig2021whenb}. Research for this work found no evidence that other Idaho counties have adopted more sophisticated algorithmic risk assessment tools, but it is unclear whether this is a result of H118 --- and the use of algorithms in the U.S.\ criminal justice system have often been hidden (see, e.g., \cite{electronicprivacyinformationcenter2020liberty,mac2021your}). A national civil society organization cited the law in a successful open records request for information related to Idaho's use of pretrial risk assessment tools \cite{electronicprivacyinformationcenter2019documents}, but Idaho civil society organizations consulted during research have not maintained active efforts centered solely on pretrial risk assessment algorithms.

\subsubsection{Discussion and other policy examples}
\label{sec:policy-uses/transparency/discussion}

Transparency requirements can be a powerful tool for algorithmic governance that does not require extensive technical expertise for policymakers to create or implement. Like legislators in most other state legislative bodies, lawmakers in Idaho have little technical subject-matter background; H118 is the only significant algorithm-related bill passed by the Idaho Legislature through the current date \cite{nationalconferenceofstatelegislatures2022legislation}. By contrast to policy proposals that attempt to create detailed rules for the use of algorithmic systems, the Idaho bill's transparency focus offered a policy solution that did not require legislators or the state government to have, develop, or hire experience with AI. However, by relying on researchers and civil society to translate the transparency it mandates into meaningful accountability, policymakers relying solely on transparency requirements effectively shift the burden of oversight from government to third parties. In the accountability framework of Metcalf et al.\ described in Section \ref{sec:background}, these efforts restrict the actors, fora, and consequences that can produce accountability \cite{metcalf2021algorithmic}.

Just as the amended Idaho bill's focus on transparency as a policy tool allowed it to bridge political disagreements, other proposals for regulating algorithmic systems have attempted to use transparency in hopes of generating information that will allow the creation of more effective regulation, or to incentivize changes in the way that algorithms are used. For example, the EU's Digital Services Act requires large online platforms to make certain data available to researchers \cite{europeancommission2022digital}, and legislative proposals for platform governance in the United States have attempted to bridge partisan gaps by adopting similar transparency requirements \cite{nonnecke2022eu}. The draft EU AI Act imposes transparency requirements on even AI systems deemed ``minimal risk'' \cite{veale2021demystifying}. Recent legislation in China created an ``algorithmic registry''  to which companies must submit detailed information about their use of algorithms, though much of this information is not intended for public consumption \cite{sheehan2022what}.

\section{Conclusion and recommendations}
\label{sec:conclusion-recommendations}

The case studies described in this paper span a spectrum from precise rules that require specific types of explanations to nonspecific requirements focused on broad forms of transparency. Each position on this spectrum offers unique benefits and drawbacks for policy. Explanation techniques developed by the machine research community --- what has been defined as ``a system [that] delivers or contains accompanying evidence or reason(s) for outputs and/or processes'' \cite{phillips2021four} (see \cite{gilpin2019explaining,barredoarrieta2020explainable} for reviews) --- allow policymakers to develop requirements that are tailored to specific audiences, impacts, or aspects of complex models or datasets. However, as this paper argues, explanation techniques developed by the research community are often too complex, too uncertain, and too restricted to satisfy the constraints that policymakers and the law operate under in practice. As a result, explanation is limited in its ability to enable meaningful public policy solutions to algorithmic harms.

Meanwhile, broad and nonspecific transparency requirements are often more approachable for policymakers in practice, even as they might be less useful or comprehensible to some audiences. Many methods proposed for increasing the accountability of algorithms, such as model cards and datasheets \cite{mitchell2019model,bender2018data,gebru2021datasheets}, audits \cite{raji2020closing}, and impact assessments \cite{moss2021assembling,metaxa2022using}, utilize a mix of broad transparency measures and technical algorithmic explanations.

The limited explicit use of explanation in policy is not only due to a lack of awareness among policymakers; the case studies in Section \ref{sec:policy-uses} illustrate several more fundamental limitations. Tools for explanation are often complex, making them unrealistic for overburdened and understaffed regulators to meaningfully evaluate. They are uncertain, often unsuitable for the kinds of binary, defensible tests that produce regulatory certainty. And they are restricted, illuminating the operation of only a small piece of complex algorithmic or socio-technical systems. These limitations do not prevent explanation from being useful for developers, but taken together, they suggest important areas where further research can make explanation more useful for policymakers.

By contrast, broader forms of transparency are much more readily accessible to policymakers. Disclosure requirements, such as Idaho HB 118 (Section \ref{sec:policy-uses/transparency}), might be burdensome for developers but are much more tractable for policymakers to impose. Inherently interpretable models (see \cite{lipton2018mythos,rudin2019stop}) are frequently used in some highly regulated industries (see Section \ref{sec:policy-uses/regulatory-approval/case-study}). Process requirements on documentation and risk management (see Section \ref{sec:policy-uses/self-regulation/discussion}) can incentivize model developers to use explanation techniques in ways that reduce fairness and safety risks.

\subsection{Recommendations for researchers}
\label{sec:discussion/takeaways}

\subsubsection{Contextualize new research with broader sets of stakeholders and broader notions of transparency and accountability.}
\label{sec:discussion/recommendations-researchers/broaden-aperture}

Literature often calls for the development of explanation tools that are useful to audiences broader than researchers and engineers \cite{miller2017explainable,tomsett2018interpretable,hoffman2021stakeholder,liao2020questioning}. Some work calls for the explicit inclusion of regulatory bodies and institutions of accountability as stakeholders \cite{naiseh2020personalising,hoffman2021stakeholder,amarasinghe2021explainable}. In interviews conducted for our case studies, we found that policymakers were very supportive of transparency and understandability in general, but that many had limited awareness of explanation as conceived in the machine learning literature. In part, this suggests an opportunity to educate policymakers. But this paper (and other literature reviewed in Section \ref{sec:background}) suggests fundamental limitations of current explainability tools that researchers might mitigate by better situating their work in the broader context of transparency and accountability. Starting points for this type of work might be the development of explainability metrics that better take policymaker constraints into account \cite{doshi-velez2017rigorous,lipton2018mythos,chen2021seven}, or literature on social scientific foundations for interpretability and explainability \cite{miller2019explanation,broniatowski2021psychological,lombrozo2006structure}. 

\subsubsection{Develop policy-informed standards for explanation.}
\label{sec:discussion/recommendations-researchers/standards}

One challenge for policymakers seeking to craft policy based on explanation is that the definition of precise requirements for explanations can limit developers' flexibility to apply the best solution to an underlying problem, especially as technology evolves. One potential strategy to mitigate these problems is the development of a technical standards for explanation. Such standards could group similar applications and use cases together into several concrete sets of requirements. This would allow regulators lacking deep technical expertise to select a set of explanation requirements that satisfy the constraints of their use case. At the same time, the definition of a small collection of unified requirements could guide research and innovation, avoiding the type of siloed application-specific development that has accompanied algorithmic systems subject to ECOA. Efforts to develop broad standards for explanation have begun in IEEE \cite{ieeestandardsassociation2021p2976,ieeestandardsassociation202270012021} and ISO \cite{internationalstandardsorganization2022iso}.

Such standards could answer calls to define a unified vocabulary for interpretability methods \cite{doshi-velez2017rigorous,arya2019one,mueller2021principles}, and could help meet recurring concerns that trade secret protections inherent to third party (or ``vendor'') models could ``shield'' models and data sources from meaningful oversight and accountability \cite{chen2021seven}. This concern was particularly evident in the Colorado insurance case study (Section \ref{sec:policy-uses/regulatory-approval}), where the regulator worried that insurers would not have a complete understanding of third party models, and the Idaho pretrial risk assessment case study (Section \ref{sec:policy-uses/transparency}), where concerns about opaque models developed by third parties drove policymaker support for legislation.

\subsection{Recommendations for policymakers}
\label{sec:discussion/recommendations-policymakers}

\subsubsection{Policy misconceptions about explainability.}
\label{sec:discussion/recommendations-policymakers/choosing}

Explanation provides just one tool for achieving broader goals of transparency and accountability. Model and data disclosure requirements \cite{mitchell2019model,bender2018data,gebru2021datasheets}, mandatory impact assessments \cite{moss2021assembling,metaxa2022using} or audits \cite{raji2020closing}, and modifications to liability law \cite{ecexpertgroup2019liability,wendehorst2022ai} all provide other means to achieve these goals.

A first question that can guide policymakers in choosing a policy approach to accountability is whether regulated entities have natural incentives to operate in good faith, or whether an adversarial approach is more appropriate. Explanation is useful on both ends of this spectrum, but in different ways. When regulators largely presume good faith, best-practice guidelines for explanation can help under-resourced companies use algorithms more responsibly. When regulators decide a more adversarial stance is appropriate, explanations can help document and justify how important development decisions were made, enabling overseers to challenge modeling choices and evaluate compliance.

Policymakers often see explainability as promoting greater user trust in algorithmic systems \cite{oecd2020recommendation,unesco2021recommendation,jobin2019global}. While explanation indeed provides one method for achieving greater accountability, trust in automated systems depends on a multitude of factors \cite{oshaughnessy2022what}, and explanation alone is insufficient to achieve trust or uptake \cite{hall2019introduction}. Moreover, transparency falls short in promoting accountability when the victims of algorithmic harms lack the financial or technical resources to interpret explanations and pursue recourse \cite{jones2023ai}. Policymakers should view explainability and understandability as two potential tools --- among many others --- to increase accountability \cite{metcalf2021algorithmic}, not simply as a goal in and of itself.

\subsubsection{Clarify objectives for transparency and explanation.}
\label{sec:discussion/recommendations-policymakers/clarify-objectives}

When explanation is used as part of a policy requirement --- either as a technical requirement or as part of a required governance process --- its objective should be clearly described. Requirements that systems be ``explainable'' should clearly define a target audience; explanations that help developers improve models are very different from explanations that provide consumers with a sense of how a system works or explanations that enable accountability from oversight bodies \cite{miller2017explainable,tomsett2018interpretable,hoffman2021stakeholder,liao2020questioning}. A specific goal is often helpful to target these explanations. For example, \cite{hoffman2021stakeholder} suggests that explanations can be particularly helpful for conveying a sense of when a model might fail or mislead --- not simply a general (and potentially incomplete) sense of ``how it works.'' Requirements or incentives for explanation should specify whether it is just the model itself that should be explained, or whether aspects of their datasets should also be transparent \cite{bender2018data}. Moreover, explanations themselves can mislead by providing users a false sense of assurance that they understand how a complex system works \cite{aivodji2019fairwashing,lakkaraju2020how,poursabzi-sangdeh2021manipulating}.

\subsubsection{Regulation can promote responsible conduct and serve business interests.}
\label{sec:discussion/recommendations-policymakers/regulation-outcomes}

Norms for transparency and explanation may evolve naturally. Companies and open-source efforts have created toolboxes and best practice guides that may help coalesce tools \cite{arya2019one}, and courts may define a common law for information that is required or expected to be provided in certain contexts \cite{deeks2019judicial,cuellar2019common}. But regulation can help guide this natural evolution, promoting responsible industry conduct while also serving business interests. Innovation often follows guardrails imposed by fast-moving regulation --- what has dubbed the ``Brussels Effect'' \cite{bradford2020brussels,siegmann2022brussels}. For example, the requirements laid out in the U.S.-centric SR 11-7 (Section \ref{sec:policy-uses/self-regulation}) have impacted global banks \cite{crespo2017evolution,sherif2016beginning} as regulators model new rules on existing guidance and developers broaden their customer base by designing algorithms to comply with global regulation. Requirements defined by ECOA in the 1970s still drive how explanation tools are used in certain industries. By carefully defining regulation that encourages the intelligent use of explanation, regulators can answer Deeks' call for ``greater involvement of public actors in shaping [explainability], which to date has largely been left in private hands'' \cite{deeks2019judicial}.

\begin{acks}
The author is grateful to Aubra Anthony, Marissa Connor, Kion Fallah, Lucy Tournas, George Perkovich, Hadrien Pouget, and Matt Sheehan for helpful discussions and feedback, and to interviewees consulted during case study research for their generosity and insight. All errors are the responsibility of the author.
\end{acks}

\bibliographystyle{ACM-Reference-Format}
\bibliography{moshaughnessy-policy.bib}


\begin{thebibliography}{154}


\ifx \showCODEN    \undefined \def \showCODEN     #1{\unskip}     \fi
\ifx \showDOI      \undefined \def \showDOI       #1{#1}\fi
\ifx \showISBNx    \undefined \def \showISBNx     #1{\unskip}     \fi
\ifx \showISBNxiii \undefined \def \showISBNxiii  #1{\unskip}     \fi
\ifx \showISSN     \undefined \def \showISSN      #1{\unskip}     \fi
\ifx \showLCCN     \undefined \def \showLCCN      #1{\unskip}     \fi
\ifx \shownote     \undefined \def \shownote      #1{#1}          \fi
\ifx \showarticletitle \undefined \def \showarticletitle #1{#1}   \fi
\ifx \showURL      \undefined \def \showURL       {\relax}        \fi
\providecommand\bibfield[2]{#2}
\providecommand\bibinfo[2]{#2}
\providecommand\natexlab[1]{#1}
\providecommand\showeprint[2][]{arXiv:#2}

\bibitem[197(1976)]%
        {1976ecoa}
 \bibinfo{year}{1976}\natexlab{}.
\newblock \bibinfo{title}{{{ECOA Amendments}} of 1976}.
\newblock
\newblock


\bibitem[202(2021)]%
        {2021remarks}
 \bibinfo{year}{2021}\natexlab{}.
\newblock \bibinfo{title}{Remarks of {{Director Rohit Chopra}} at a {{Joint
  DOJ}}, {{CFPB}}, and {{OCC Press Conference}} on the {{Trustmark National
  Bank Enforcement Action}}}.
\newblock
\newblock
\urldef\tempurl%
\url{https://www.consumerfinance.gov/about-us/newsroom/remarks-of-director-rohit-chopra-at-a-joint-doj-cfpb-and-occ-press-conference-on-the-trustmark-national-bank-enforcement-action/}
\showURL{%
\tempurl}


\bibitem[202(2022)]%
        {2022cfpb}
 \bibinfo{year}{2022}\natexlab{}.
\newblock \bibinfo{title}{{{CFPB Launches New Effort}} to {{Promote
  Competition}} and {{Innovation}} in {{Consumer Finance}}}.
\newblock
\newblock
\urldef\tempurl%
\url{https://www.consumerfinance.gov/about-us/newsroom/cfpb-lauches-new-effort-to-promote-competition-and-innovation-in-consumer-finance/}
\showURL{%
\tempurl}


\bibitem[94th Congress(1976)]%
        {94thcongress1976senate}
\bibfield{author}{\bibinfo{person}{94th Congress}.}
  \bibinfo{year}{1976}\natexlab{}.
\newblock \bibinfo{booktitle}{\emph{Senate {{Report}} 94-589}}.
\newblock \bibinfo{type}{{T}echnical {R}eport}. \bibinfo{institution}{{94th
  Congress, 2nd Session}}.
\newblock


\bibitem[Abraham and Schwarcz(2022)]%
        {abraham2022limits}
\bibfield{author}{\bibinfo{person}{Kenneth~S. Abraham} {and}
  \bibinfo{person}{Daniel Schwarcz}.} \bibinfo{year}{2022}\natexlab{}.
\newblock \showarticletitle{The {{Limits}} of {{Regulation}} by {{Insurance}}}.
\newblock \bibinfo{journal}{\emph{Indiana Law Journal}}  \bibinfo{volume}{98}
  (\bibinfo{year}{2022}).
\newblock


\bibitem[Aivodji et~al\mbox{.}(2019)]%
        {aivodji2019fairwashing}
\bibfield{author}{\bibinfo{person}{Ulrich Aivodji}, \bibinfo{person}{Hiromi
  Arai}, \bibinfo{person}{Olivier Fortineau}, \bibinfo{person}{S{\'e}bastien
  Gambs}, \bibinfo{person}{Satoshi Hara}, {and} \bibinfo{person}{Alain Tapp}.}
  \bibinfo{year}{2019}\natexlab{}.
\newblock \showarticletitle{Fairwashing: The Risk of Rationalization}. In
  \bibinfo{booktitle}{\emph{Proceedings of the 36th {{International
  Conference}} on {{Machine Learning}}}}. \bibinfo{publisher}{{PMLR}},
  \bibinfo{pages}{161--170}.
\newblock
\showISSN{2640-3498}
\urldef\tempurl%
\url{https://proceedings.mlr.press/v97/aivodji19a.html}
\showURL{%
\tempurl}


\bibitem[Akinwumi et~al\mbox{.}(2021)]%
        {akinwumi2021ai}
\bibfield{author}{\bibinfo{person}{Michael Akinwumi}, \bibinfo{person}{John
  Merrill}, \bibinfo{person}{Lisa Rice}, \bibinfo{person}{Kareem Saleh}, {and}
  \bibinfo{person}{Maureen Yap}.} \bibinfo{year}{2021}\natexlab{}.
\newblock \bibinfo{booktitle}{\emph{An {{AI}} Fair Lending Policy Agenda for
  the Federal Financial Regulators}}.
\newblock \bibinfo{type}{{T}echnical {R}eport}.
  \bibinfo{institution}{{Brookings Institution}}.
\newblock
\urldef\tempurl%
\url{https://www.brookings.edu/research/an-ai-fair-lending-policy-agenda-for-the-federal-financial-regulators/}
\showURL{%
\tempurl}


\bibitem[Amarasinghe et~al\mbox{.}(2021)]%
        {amarasinghe2021explainable}
\bibfield{author}{\bibinfo{person}{Kasun Amarasinghe}, \bibinfo{person}{Kit
  Rodolfa}, \bibinfo{person}{Hemank Lamba}, {and} \bibinfo{person}{Rayid
  Ghani}.} \bibinfo{year}{2021}\natexlab{}.
\newblock \bibinfo{title}{Explainable {{Machine Learning}} for {{Public
  Policy}}: {{Use Cases}}, {{Gaps}}, and {{Research Directions}}}.
\newblock
\newblock
\showeprint[arxiv]{2010.14374}~[cs]
\urldef\tempurl%
\url{http://arxiv.org/abs/2010.14374}
\showURL{%
\tempurl}


\bibitem[Ammermann(2013)]%
        {ammermann2013adverse}
\bibfield{author}{\bibinfo{person}{Sarah Ammermann}.}
  \bibinfo{year}{2013}\natexlab{}.
\newblock \showarticletitle{Adverse {{Action Notice Requirements Under}} the
  {{ECOA}} and the {{FCRA}}}.
\newblock \bibinfo{journal}{\emph{Consumer Compliance Outlook}}
  (\bibinfo{year}{2013}).
\newblock
\urldef\tempurl%
\url{https://www.consumercomplianceoutlook.org/2013/second-quarter/adverse-action-notice-requirements-under-ecoa-fcra/}
\showURL{%
\tempurl}


\bibitem[Andrews(2008)]%
        {andrews2008greenspan}
\bibfield{author}{\bibinfo{person}{Edmund~L. Andrews}.}
  \bibinfo{year}{2008}\natexlab{}.
\newblock \showarticletitle{Greenspan {{Concedes Error}} on {{Regulation}}}.
\newblock \bibinfo{journal}{\emph{The New York Times}} (\bibinfo{date}{Oct.}
  \bibinfo{year}{2008}).
\newblock
\showISSN{0362-4331}
\urldef\tempurl%
\url{https://www.nytimes.com/2008/10/24/business/economy/24panel.html}
\showURL{%
\tempurl}


\bibitem[Angwin et~al\mbox{.}(2017)]%
        {angwin2017minority}
\bibfield{author}{\bibinfo{person}{Julia Angwin}, \bibinfo{person}{Jeff
  Larson}, \bibinfo{person}{Lauren Kirchner}, {and} \bibinfo{person}{Surya
  Mattu}.} \bibinfo{year}{2017}\natexlab{}.
\newblock \bibinfo{title}{Minority {{Neighborhoods Pay Higher Car Insurance
  Premiums Than White Areas With}} the {{Same Risk}}}.
\newblock
\newblock
\urldef\tempurl%
\url{https://www.propublica.org/article/minority-neighborhoods-higher-car-insurance-premiums-white-areas-same-risk}
\showURL{%
\tempurl}


\bibitem[Angwin et~al\mbox{.}(2016)]%
        {angwin2016machine}
\bibfield{author}{\bibinfo{person}{Julia Angwin}, \bibinfo{person}{Jeff
  Larson}, \bibinfo{person}{Surya Mattu}, {and} \bibinfo{person}{Lauren
  Kirchner}.} \bibinfo{year}{2016}\natexlab{}.
\newblock \bibinfo{booktitle}{\emph{Machine {{Bias}}}}.
\newblock \bibinfo{type}{{T}echnical {R}eport}.
  \bibinfo{institution}{{ProPublica}}.
\newblock
\urldef\tempurl%
\url{https://www.propublica.org/article/machine-bias-risk-assessments-in-criminal-sentencing}
\showURL{%
\tempurl}


\bibitem[Armstrong(2019)]%
        {armstrong2019board}
\bibfield{author}{\bibinfo{person}{Thomas~H. Armstrong}.}
  \bibinfo{year}{2019}\natexlab{}.
\newblock \bibinfo{title}{Board of {{Governors}} of the {{Federal Reserve
  System}} - {{Applicability}} of the {{Congressional Review Act}} to
  {{Supervision}} and {{Regulation Letter}} 11-7}.
\newblock
\newblock
\urldef\tempurl%
\url{https://www.gao.gov/assets/b-331324.pdf}
\showURL{%
\tempurl}


\bibitem[Aronowitz(2012)]%
        {aronowitz2012rise}
\bibfield{author}{\bibinfo{person}{Robert~A. Aronowitz}.}
  \bibinfo{year}{2012}\natexlab{}.
\newblock \showarticletitle{The {{Rise}} and {{Fall}} of the {{Lyme Disease
  Vaccines}}: {{A Cautionary Tale}} for {{Risk Interventions}} in {{American
  Medicine}} and {{Public Health}}}.
\newblock \bibinfo{journal}{\emph{Milbank Quarterly}} \bibinfo{volume}{90},
  \bibinfo{number}{2} (\bibinfo{date}{June} \bibinfo{year}{2012}),
  \bibinfo{pages}{250--277}.
\newblock
\showISSN{0887378X}
\urldef\tempurl%
\url{https://doi.org/10.1111/j.1468-0009.2012.00663.x}
\showDOI{\tempurl}


\bibitem[Arya et~al\mbox{.}(2019)]%
        {arya2019one}
\bibfield{author}{\bibinfo{person}{Vijay Arya}, \bibinfo{person}{Rachel K.~E.
  Bellamy}, \bibinfo{person}{Pin-Yu Chen}, \bibinfo{person}{Amit Dhurandhar},
  \bibinfo{person}{Michael Hind}, \bibinfo{person}{Samuel~C. Hoffman},
  \bibinfo{person}{Stephanie Houde}, \bibinfo{person}{Q.~Vera Liao},
  \bibinfo{person}{Ronny Luss}, \bibinfo{person}{Aleksandra Mojsilovi{\'c}},
  \bibinfo{person}{Sami Mourad}, \bibinfo{person}{Pablo Pedemonte},
  \bibinfo{person}{Ramya Raghavendra}, \bibinfo{person}{John Richards},
  \bibinfo{person}{Prasanna Sattigeri}, \bibinfo{person}{Karthikeyan
  Shanmugam}, \bibinfo{person}{Moninder Singh}, \bibinfo{person}{Kush~R.
  Varshney}, \bibinfo{person}{Dennis Wei}, {and} \bibinfo{person}{Yunfeng
  Zhang}.} \bibinfo{year}{2019}\natexlab{}.
\newblock \showarticletitle{One {{Explanation Does Not Fit All}}: {{A Toolkit}}
  and {{Taxonomy}} of {{AI Explainability Techniques}}}.
\newblock  (\bibinfo{year}{2019}).
\newblock
\urldef\tempurl%
\url{https://doi.org/10.48550/ARXIV.1909.03012}
\showDOI{\tempurl}


\bibitem[Assembly(2019)]%
        {illinoisgeneralassembly2019artificial}
\bibfield{author}{\bibinfo{person}{Illinois~General Assembly}.}
  \bibinfo{year}{2019}\natexlab{}.
\newblock \bibinfo{title}{Artificial {{Intelligence Video Interview Act}}}.
\newblock
\newblock
\urldef\tempurl%
\url{https://www.ilga.gov/legislation/ilcs/ilcs3.asp?ActID=4015&ChapterID=68}
\showURL{%
\tempurl}


\bibitem[Association(2021)]%
        {ieeestandardsassociation2021p2976}
\bibfield{author}{\bibinfo{person}{IEEE~Standards Association}.}
  \bibinfo{year}{2021}\natexlab{}.
\newblock \bibinfo{title}{P2976: {{Standard}} for {{XAI}} \textendash{}
  {{eXplainable Artificial Intelligence}} - for {{Achieving Clarity}} and
  {{Interoperability}} of {{AI Systems Design}}}.
\newblock
\newblock
\urldef\tempurl%
\url{https://standards.ieee.org/ieee/2976/10522/}
\showURL{%
\tempurl}


\bibitem[Association(2022)]%
        {ieeestandardsassociation202270012021}
\bibfield{author}{\bibinfo{person}{IEEE~Standards Association}.}
  \bibinfo{year}{2022}\natexlab{}.
\newblock \bibinfo{booktitle}{\emph{7001-2021 - {{IEEE Standard}} for
  {{Transparency}} of {{Autonomous Systems}}}}.
\newblock \bibinfo{type}{{T}echnical {R}eport}. \bibinfo{institution}{{IEEE}}.
\newblock
\showISBNx{9781504483117}
\urldef\tempurl%
\url{https://doi.org/10.1109/IEEESTD.2022.9726144}
\showDOI{\tempurl}


\bibitem[Barocas and Selbst(2016)]%
        {barocas2016big}
\bibfield{author}{\bibinfo{person}{Solon Barocas} {and} \bibinfo{person}{Andrew
  Selbst}.} \bibinfo{year}{2016}\natexlab{}.
\newblock \showarticletitle{Big {{Data}}'s {{Disparate Impact}}}.
\newblock \bibinfo{journal}{\emph{California Law Review}}
  \bibinfo{volume}{104} (\bibinfo{year}{2016}), \bibinfo{pages}{671}.
\newblock
\urldef\tempurl%
\url{https://doi.org/10.15779/Z38BG31}
\showDOI{\tempurl}


\bibitem[Barredo~Arrieta et~al\mbox{.}(2020)]%
        {barredoarrieta2020explainable}
\bibfield{author}{\bibinfo{person}{Alejandro Barredo~Arrieta},
  \bibinfo{person}{Natalia {D{\'i}az-Rodr{\'i}guez}}, \bibinfo{person}{Javier
  Del~Ser}, \bibinfo{person}{Adrien Bennetot}, \bibinfo{person}{Siham Tabik},
  \bibinfo{person}{Alberto Barbado}, \bibinfo{person}{Salvador Garcia},
  \bibinfo{person}{Sergio {Gil-Lopez}}, \bibinfo{person}{Daniel Molina},
  \bibinfo{person}{Richard Benjamins}, \bibinfo{person}{Raja Chatila}, {and}
  \bibinfo{person}{Francisco Herrera}.} \bibinfo{year}{2020}\natexlab{}.
\newblock \showarticletitle{Explainable {{Artificial Intelligence}} ({{XAI}}):
  {{Concepts}}, Taxonomies, Opportunities and Challenges toward Responsible
  {{AI}}}.
\newblock \bibinfo{journal}{\emph{Information Fusion}}  \bibinfo{volume}{58}
  (\bibinfo{date}{June} \bibinfo{year}{2020}), \bibinfo{pages}{82--115}.
\newblock
\showISSN{15662535}
\urldef\tempurl%
\url{https://doi.org/10.1016/j.inffus.2019.12.012}
\showDOI{\tempurl}


\bibitem[Bathaee(2018)]%
        {bathaee2018artificial}
\bibfield{author}{\bibinfo{person}{Yavar Bathaee}.}
  \bibinfo{year}{2018}\natexlab{}.
\newblock \showarticletitle{The {{Artificial Intelligence Black Box}} and the
  {{Failure}} of {{Intent}} and {{Causation}}}.
\newblock \bibinfo{journal}{\emph{Harvard Journal of Law \& Technology}}
  \bibinfo{volume}{31}, \bibinfo{number}{2} (\bibinfo{year}{2018}).
\newblock


\bibitem[Bender and Friedman(2018)]%
        {bender2018data}
\bibfield{author}{\bibinfo{person}{Emily~M. Bender} {and}
  \bibinfo{person}{Batya Friedman}.} \bibinfo{year}{2018}\natexlab{}.
\newblock \showarticletitle{Data {{Statements}} for {{Natural Language
  Processing}}: {{Toward Mitigating System Bias}} and {{Enabling Better
  Science}}}.
\newblock \bibinfo{journal}{\emph{Transactions of the Association for
  Computational Linguistics}}  \bibinfo{volume}{6} (\bibinfo{date}{Dec.}
  \bibinfo{year}{2018}), \bibinfo{pages}{587--604}.
\newblock
\showISSN{2307-387X}
\urldef\tempurl%
\url{https://doi.org/10.1162/tacl_a_00041}
\showDOI{\tempurl}


\bibitem[Bhatt et~al\mbox{.}(2020a)]%
        {bhatt2020machine}
\bibfield{author}{\bibinfo{person}{Umang Bhatt}, \bibinfo{person}{McKane
  Andrus}, \bibinfo{person}{Adrian Weller}, {and} \bibinfo{person}{Alice
  Xiang}.} \bibinfo{year}{2020}\natexlab{a}.
\newblock \bibinfo{title}{Machine {{Learning Explainability}} for {{External
  Stakeholders}}}.
\newblock
\newblock
\urldef\tempurl%
\url{http://arxiv.org/abs/2007.05408}
\showURL{%
\tempurl}


\bibitem[Bhatt et~al\mbox{.}(2020b)]%
        {bhatt2020explainable}
\bibfield{author}{\bibinfo{person}{Umang Bhatt}, \bibinfo{person}{Alice Xiang},
  \bibinfo{person}{Shubham Sharma}, \bibinfo{person}{Adrian Weller},
  \bibinfo{person}{Ankur Taly}, \bibinfo{person}{Yunhan Jia},
  \bibinfo{person}{Joydeep Ghosh}, \bibinfo{person}{Ruchir Puri},
  \bibinfo{person}{Jos{\'e} M.~F. Moura}, {and} \bibinfo{person}{Peter
  Eckersley}.} \bibinfo{year}{2020}\natexlab{b}.
\newblock \showarticletitle{Explainable Machine Learning in Deployment}. In
  \bibinfo{booktitle}{\emph{Proceedings of the 2020 {{Conference}} on
  {{Fairness}}, {{Accountability}}, and {{Transparency}}}}.
  \bibinfo{publisher}{{ACM}}, \bibinfo{address}{{Barcelona Spain}},
  \bibinfo{pages}{648--657}.
\newblock
\showISBNx{978-1-4503-6936-7}
\urldef\tempurl%
\url{https://doi.org/10.1145/3351095.3375624}
\showDOI{\tempurl}


\bibitem[Black(2021)]%
        {black2021insurers}
\bibfield{author}{\bibinfo{person}{Ann~Young Black}.}
  \bibinfo{year}{2021}\natexlab{}.
\newblock \bibinfo{title}{Insurers {{Need}} to {{Do Their Homework}}:
  {{Review}} of the {{Use}} of {{Data}}, {{Algorithms}}, and {{Predictive
  Models}}}.
\newblock
\newblock
\urldef\tempurl%
\url{https://www.carltonfields.com/insights/expect-focus/2021/review-use-of-data-algorithms-predictive-models}
\showURL{%
\tempurl}


\bibitem[{Board of Governors of the Federal Reserve System {and} Office of the
  Comptroller of the Currency}(2011)]%
        {federalreserve2011guidance}
\bibfield{author}{\bibinfo{person}{{Board of Governors of the Federal Reserve
  System {and} Office of the Comptroller of the Currency}}.}
  \bibinfo{year}{2011}\natexlab{}.
\newblock \bibinfo{booktitle}{\emph{{{SR}} 11-7: {{Guidance}} on {{Model Risk
  Management}}}}.
\newblock \bibinfo{type}{{T}echnical {R}eport}.
\newblock
\urldef\tempurl%
\url{https://www.federalreserve.gov/supervisionreg/srletters/sr1107.htm}
\showURL{%
\tempurl}


\bibitem[Bogen et~al\mbox{.}(2020)]%
        {bogen2020awareness}
\bibfield{author}{\bibinfo{person}{Miranda Bogen}, \bibinfo{person}{Aaron
  Rieke}, {and} \bibinfo{person}{Shazeda Ahmed}.}
  \bibinfo{year}{2020}\natexlab{}.
\newblock \showarticletitle{Awareness in Practice: Tensions in Access to
  Sensitive Attribute Data for Antidiscrimination}. In
  \bibinfo{booktitle}{\emph{Proceedings of the 2020 {{Conference}} on
  {{Fairness}}, {{Accountability}}, and {{Transparency}}}}.
  \bibinfo{publisher}{{ACM}}, \bibinfo{address}{{Barcelona Spain}},
  \bibinfo{pages}{492--500}.
\newblock
\showISBNx{978-1-4503-6936-7}
\urldef\tempurl%
\url{https://doi.org/10.1145/3351095.3372877}
\showDOI{\tempurl}


\bibitem[Bradford(2020)]%
        {bradford2020brussels}
\bibfield{author}{\bibinfo{person}{Anu Bradford}.}
  \bibinfo{year}{2020}\natexlab{}.
\newblock \bibinfo{booktitle}{\emph{The {{Brussels}} Effect: How the {{European
  Union}} Rules the World}}.
\newblock \bibinfo{publisher}{{Oxford University Press}},
  \bibinfo{address}{{New York}}.
\newblock
\showISBNx{978-0-19-008861-3 978-0-19-008860-6 978-0-19-008859-0}
\showLCCN{K590.5}


\bibitem[Bradley(2016)]%
        {bradley2016state}
\bibfield{author}{\bibinfo{person}{Ann~Walsh Bradley}.}
  \bibinfo{year}{2016}\natexlab{}.
\newblock \bibinfo{title}{State v. {{Loomis}}}.
\newblock
\newblock


\bibitem[Bradley et~al\mbox{.}(2021)]%
        {bradley2021national}
\bibfield{author}{\bibinfo{person}{Charles Bradley}, \bibinfo{person}{Richard
  Wingfield}, {and} \bibinfo{person}{Megan Metzger}.}
  \bibinfo{year}{2021}\natexlab{}.
\newblock \bibinfo{booktitle}{\emph{National {{Artificial Intelligence
  Strategies}} and {{Human Rights}}: {{A Review}}, {{Second Edition}}}}.
\newblock \bibinfo{type}{{T}echnical {R}eport}. \bibinfo{institution}{{Stanford
  Cyber Policy Center}}.
\newblock


\bibitem[Broniatowski(2021)]%
        {broniatowski2021psychological}
\bibfield{author}{\bibinfo{person}{David~A. Broniatowski}.}
  \bibinfo{year}{2021}\natexlab{}.
\newblock \bibinfo{booktitle}{\emph{Psychological {{Foundations}} of
  {{Explainability}} and {{Interpretability}} in {{Artificial Intelligence}}}}.
\newblock \bibinfo{type}{{T}echnical {R}eport} NISTIR 8367.
  \bibinfo{institution}{{U.S. National Institute for Standards and
  Technology}}.
\newblock
\urldef\tempurl%
\url{https://nvlpubs.nist.gov/nistpubs/ir/2021/NIST.IR.8367.pdf}
\showURL{%
\tempurl}


\bibitem[Brotcke(2022)]%
        {brotcke2022time}
\bibfield{author}{\bibinfo{person}{Liming Brotcke}.}
  \bibinfo{year}{2022}\natexlab{}.
\newblock \showarticletitle{Time to {{Assess Bias}} in {{Machine Learning
  Models}} for {{Credit Decisions}}}.
\newblock \bibinfo{journal}{\emph{Journal of Risk and Financial Management}}
  \bibinfo{volume}{15}, \bibinfo{number}{4} (\bibinfo{date}{April}
  \bibinfo{year}{2022}), \bibinfo{pages}{165}.
\newblock
\showISSN{1911-8074}
\urldef\tempurl%
\url{https://doi.org/10.3390/jrfm15040165}
\showDOI{\tempurl}


\bibitem[Buckner et~al\mbox{.}(2021)]%
        {buckner2021concerning}
\bibfield{author}{\bibinfo{person}{Janet Buckner}, \bibinfo{person}{Rhonda
  Fields}, \bibinfo{person}{Dominick Moreno}, {and} \bibinfo{person}{Tammy
  Story}.} \bibinfo{year}{2021}\natexlab{}.
\newblock \bibinfo{title}{Concerning {{Protecting Consumers}} from {{Unfair
  Discrimination}} in {{Insurance Practices}}}.
\newblock
\newblock
\urldef\tempurl%
\url{https://www.leg.colorado.gov/sites/default/files/2021a_169_signed.pdf}
\showURL{%
\tempurl}


\bibitem[Buolamwini and Gebru(2018)]%
        {buolamwini2018gender}
\bibfield{author}{\bibinfo{person}{Joy Buolamwini} {and}
  \bibinfo{person}{Timnit Gebru}.} \bibinfo{year}{2018}\natexlab{}.
\newblock \showarticletitle{Gender {{Shades}}: {{Intersectional Accuracy
  Disparities}} in {{Commercial Gender Classificiation}}}. In
  \bibinfo{booktitle}{\emph{Proc. {{Machine Learning Research}}}}.
  \bibinfo{address}{{New York, NY, USA}}.
\newblock


\bibitem[Bureau(2018)]%
        {consumerfinancialprotectionbureau201812}
\bibfield{author}{\bibinfo{person}{Consumer Financial~Protection Bureau}.}
  \bibinfo{year}{2018}\natexlab{}.
\newblock \bibinfo{booktitle}{\emph{12 {{CFR Part}} 1002 - {{Equal Credit
  Opportunity Act}} ({{Regulation B}})}}.
\newblock \bibinfo{type}{{T}echnical {R}eport}.
\newblock
\urldef\tempurl%
\url{https://www.consumerfinance.gov/rules-policy/regulations/1002/}
\showURL{%
\tempurl}


\bibitem[Bureau(2022)]%
        {consumerfinancialprotectionbureau2022consumer}
\bibfield{author}{\bibinfo{person}{Consumer Financial~Protection Bureau}.}
  \bibinfo{year}{2022}\natexlab{}.
\newblock \bibinfo{title}{Consumer {{Financial Protection Circular}} 2022-03}.
\newblock
\newblock
\urldef\tempurl%
\url{https://www.consumerfinance.gov/compliance/circulars/circular-2022-03-adverse-action-notification-requirements-in-connection-with-credit-decisions-based-on-complex-algorithms/#1}
\showURL{%
\tempurl}


\bibitem[Bureau(2014)]%
        {u.s.consumerfinancialprotectionbureau2014using}
\bibfield{author}{\bibinfo{person}{U.S. Consumer Financial~Protection Bureau}.}
  \bibinfo{year}{2014}\natexlab{}.
\newblock \bibinfo{booktitle}{\emph{Using Publicly Available Information to
  Proxy for Unidentified Race and Ethnicity: {{A}} Methodology and
  Assessment}}.
\newblock \bibinfo{type}{{T}echnical {R}eport}.
\newblock
\urldef\tempurl%
\url{https://files.consumerfinance.gov/f/201409_cfpb_report_proxy-methodology.pdf}
\showURL{%
\tempurl}


\bibitem[Cardi(2011)]%
        {cardi2011hidden}
\bibfield{author}{\bibinfo{person}{W.~Jonathan Cardi}.}
  \bibinfo{year}{2011}\natexlab{}.
\newblock \showarticletitle{The {{Hidden Legacy}} of {{Palsgraf}}: {{Modern
  Duty Law}} in {{Microcosm}}}.
\newblock \bibinfo{journal}{\emph{Boston University Law Review}}
  \bibinfo{volume}{91} (\bibinfo{year}{2011}), \bibinfo{pages}{1873--1913}.
\newblock


\bibitem[Center(2019)]%
        {electronicprivacyinformationcenter2019documents}
\bibfield{author}{\bibinfo{person}{Electronic Privacy~Information Center}.}
  \bibinfo{year}{2019}\natexlab{}.
\newblock \bibinfo{title}{Documents {{Obtained}} by {{EPIC}} Show {{Idaho}}'s
  {{Use}} of {{Subjective Categories}} in {{Calculating Risk}}}.
\newblock
\newblock


\bibitem[Center(2020)]%
        {electronicprivacyinformationcenter2020liberty}
\bibfield{author}{\bibinfo{person}{Electronic Privacy~Information Center}.}
  \bibinfo{year}{2020}\natexlab{}.
\newblock \bibinfo{booktitle}{\emph{Liberty at {{Risk}}: {{Pre-trial Risk
  Assessment Tools}} in the {{U}}.{{S}}.}}
\newblock \bibinfo{type}{{T}echnical {R}eport}.
\newblock
\urldef\tempurl%
\url{https://archive.epic.org/LibertyAtRiskReport.pdf}
\showURL{%
\tempurl}


\bibitem[Chaney(2019)]%
        {chaney2019criminal}
\bibfield{author}{\bibinfo{person}{Greg Chaney}.}
  \bibinfo{year}{2019}\natexlab{}.
\newblock \showarticletitle{The {{Criminal Justice System}}'s {{Algorithms Need
  Transparency}}}.
\newblock \bibinfo{journal}{\emph{Law360}} (\bibinfo{date}{March}
  \bibinfo{year}{2019}).
\newblock
\urldef\tempurl%
\url{https://www.law360.com/articles/1143086/the-criminal-justice-system-s-algorithms-need-transparency}
\showURL{%
\tempurl}


\bibitem[Chen and Storchan(2021)]%
        {chen2021seven}
\bibfield{author}{\bibinfo{person}{Jiahao Chen} {and} \bibinfo{person}{Victor
  Storchan}.} \bibinfo{year}{2021}\natexlab{}.
\newblock \showarticletitle{Seven Challenges for Harmonizing Explainability
  Requirements}. In \bibinfo{booktitle}{\emph{Proc. {{KDD Workshop}} on
  {{Machine Learning}} in {{Finance}}}}. \bibinfo{publisher}{{ACM}}.
\newblock


\bibitem[Collazo(2021)]%
        {collazo2021advanced}
\bibfield{author}{\bibinfo{person}{Dolores Collazo}.}
  \bibinfo{year}{2021}\natexlab{}.
\newblock \showarticletitle{Advanced {{Topics}} in {{Adverse Action Notices
  Under}} the {{Equal Credit Opportunity Act}}}.
\newblock \bibinfo{journal}{\emph{Consumer Compliance Outlook}}
  \bibinfo{number}{Fourth Issue} (\bibinfo{year}{2021}).
\newblock
\urldef\tempurl%
\url{https://www.consumercomplianceoutlook.org/2021/fourth-issue/advanced-topics-in-adverse-action-notices-under-the-equal-credit-opportunity-act/}
\showURL{%
\tempurl}


\bibitem[{Colorado Division of Insurance}(2022a)]%
        {coloradodivisionofinsurance2022recordinga}
\bibfield{author}{\bibinfo{person}{{Colorado Division of Insurance}}.}
  \bibinfo{year}{2022}\natexlab{a}.
\newblock \bibinfo{title}{Recording from {{April}} 12, 2022 {{Stakeholder
  Meeting}}}.
\newblock
\newblock
\urldef\tempurl%
\url{https://doi.colorado.gov/for-consumers/sb21-169-protecting-consumers-from-unfair-discrimination-in-insurance-practices}
\showURL{%
\tempurl}


\bibitem[{Colorado Division of Insurance}(2022b)]%
        {coloradodivisionofinsurance2022recording}
\bibfield{author}{\bibinfo{person}{{Colorado Division of Insurance}}.}
  \bibinfo{year}{2022}\natexlab{b}.
\newblock \bibinfo{title}{Recording from {{December}} 8, 2022 {{Stakeholder
  Meeting}}}.
\newblock
\newblock
\urldef\tempurl%
\url{https://doi.colorado.gov/for-consumers/sb21-169-protecting-consumers-from-unfair-discrimination-in-insurance-practices}
\showURL{%
\tempurl}


\bibitem[{Colorado Division of Insurance}(2022c)]%
        {coloradodivisionofinsurance2022recordingb}
\bibfield{author}{\bibinfo{person}{{Colorado Division of Insurance}}.}
  \bibinfo{year}{2022}\natexlab{c}.
\newblock \bibinfo{title}{Recording from {{July}} 8, 2022 {{Stakeholder
  Meeting}}}.
\newblock
\newblock
\urldef\tempurl%
\url{https://doi.colorado.gov/for-consumers/sb21-169-protecting-consumers-from-unfair-discrimination-in-insurance-practices}
\showURL{%
\tempurl}


\bibitem[{Colorado Division of Insurance}(2022d)]%
        {coloradodivisionofinsurance2022recordingc}
\bibfield{author}{\bibinfo{person}{{Colorado Division of Insurance}}.}
  \bibinfo{year}{2022}\natexlab{d}.
\newblock \bibinfo{title}{Recording from {{September}} 28, 2022 {{Stakeholder
  Meeting}}}.
\newblock
\newblock
\urldef\tempurl%
\url{https://doi.colorado.gov/for-consumers/sb21-169-protecting-consumers-from-unfair-discrimination-in-insurance-practices}
\showURL{%
\tempurl}


\bibitem[{Colorado Division of Insurance}(2022e)]%
        {coloradodivisionofinsurance2022slide}
\bibfield{author}{\bibinfo{person}{{Colorado Division of Insurance}}.}
  \bibinfo{year}{2022}\natexlab{e}.
\newblock \bibinfo{title}{Slide {{Deck}} from {{February}} 17, 2022
  {{Stakeholder Meeting}}}.
\newblock
\newblock
\urldef\tempurl%
\url{https://doi.colorado.gov/for-consumers/sb21-169-protecting-consumers-from-unfair-discrimination-in-insurance-practices}
\showURL{%
\tempurl}


\bibitem[Commission(2021)]%
        {europeancommission2021proposal}
\bibfield{author}{\bibinfo{person}{European Commission}.}
  \bibinfo{year}{2021}\natexlab{}.
\newblock \bibinfo{booktitle}{\emph{Proposal for a {{Regulation}} of the
  {{European Parliament}} and of the {{Council}}: {{Laying Down Harmonised}}
  Rules on {{Artificial Intelligence}} ({{Artificial Intelligence Act}}) and
  {{Amending Certain Union Legislative Acts}}}}.
\newblock \bibinfo{type}{{T}echnical {R}eport} 2021/0106 (COD).
\newblock


\bibitem[Commission(2022a)]%
        {europeancommission2022digital}
\bibfield{author}{\bibinfo{person}{European Commission}.}
  \bibinfo{year}{2022}\natexlab{a}.
\newblock \bibinfo{title}{The {{Digital Services Act}} Package}.
\newblock
\newblock
\urldef\tempurl%
\url{https://digital-strategy.ec.europa.eu/en/policies/digital-services-act-package}
\showURL{%
\tempurl}


\bibitem[Commission(2022b)]%
        {europeancommission2022draft}
\bibfield{author}{\bibinfo{person}{European Commission}.}
  \bibinfo{year}{2022}\natexlab{b}.
\newblock \bibinfo{title}{Draft Standardisation Request to the {{European
  Standardisation Organisations}} in Support of Safe and Trustworthy Artificial
  Intelligence}.
\newblock
\newblock
\urldef\tempurl%
\url{https://ec.europa.eu/docsroom/documents/52376?locale=en}
\showURL{%
\tempurl}


\bibitem[Commission(2022c)]%
        {europeancommission2022proposal}
\bibfield{author}{\bibinfo{person}{European Commission}.}
  \bibinfo{year}{2022}\natexlab{c}.
\newblock \bibinfo{title}{Proposal for a {{Directive}} on Adapting Non
  Contractual Civil Liability Rules to Artificial Intelligence}.
\newblock
\newblock
\urldef\tempurl%
\url{https://commission.europa.eu/business-economy-euro/doing-business-eu/contract-rules/digital-contracts/liability-rules-artificial-intelligence_en}
\showURL{%
\tempurl}


\bibitem[Commission(2013)]%
        {u.s.federaltradecommission2013ftc}
\bibfield{author}{\bibinfo{person}{U.S. Federal~Trade Commission}.}
  \bibinfo{year}{2013}\natexlab{}.
\newblock \bibinfo{title}{In {{FTC Study}}, {{Five Percent}} of {{Consumers Had
  Errors}} on {{Their Credit Reports That Could Result}} in {{Less Favorable
  Terms}} for {{Loans}}}.
\newblock
\newblock
\urldef\tempurl%
\url{https://www.ftc.gov/news-events/news/press-releases/2013/02/ftc-study-five-percent-consumers-had-errors-their-credit-reports-could-result-less-favorable-terms}
\showURL{%
\tempurl}


\bibitem[Congress(2022)]%
        {shenzhenmunicipalpeoplescongress2022shenzhen}
\bibfield{author}{\bibinfo{person}{Shenzhen Municipal~People's Congress}.}
  \bibinfo{year}{2022}\natexlab{}.
\newblock \bibinfo{title}{{Shenzhen Special Economic Zone Artificial
  Intelligence Industry Promotion Regulations [Chinese]}}.
\newblock
\newblock
\urldef\tempurl%
\url{https://law.pkulaw.com/chinalaw/eb370a7e0d9edd5e8ca8bb1a5fa6a5e7bdfb.html}
\showURL{%
\tempurl}


\bibitem[Corporation(2021)]%
        {randcorporation2021bayesian}
\bibfield{author}{\bibinfo{person}{RAND Corporation}.}
  \bibinfo{year}{2021}\natexlab{}.
\newblock \bibinfo{title}{Bayesian {{Indirect Surname Geocoding}} ({{BISG}})}.
\newblock
\newblock
\urldef\tempurl%
\url{https://www.rand.org/health-care/tools-methods/bisg.html}
\showURL{%
\tempurl}


\bibitem[Crespo et~al\mbox{.}(2017)]%
        {crespo2017evolution}
\bibfield{author}{\bibinfo{person}{Ignacio Crespo}, \bibinfo{person}{Pankaj
  Kumar}, \bibinfo{person}{Peter Noteboom}, {and} \bibinfo{person}{Marc
  Taymans}.} \bibinfo{year}{2017}\natexlab{}.
\newblock \bibinfo{booktitle}{\emph{The Evolution of Model Risk Management}}.
\newblock \bibinfo{type}{{T}echnical {R}eport}. \bibinfo{institution}{{McKinsey
  \& Company}}.
\newblock
\urldef\tempurl%
\url{https://www.mckinsey.com/capabilities/risk-and-resilience/our-insights/the-evolution-of-model-risk-management}
\showURL{%
\tempurl}


\bibitem[Cu{\'e}llar(2019)]%
        {cuellar2019common}
\bibfield{author}{\bibinfo{person}{Mariano-Florentino Cu{\'e}llar}.}
  \bibinfo{year}{2019}\natexlab{}.
\newblock \showarticletitle{A {{Common Law}} for the {{Age}} of {{Artificial
  Intelligence}}: {{Incremental Adjudication}}, {{Institutions}}, and
  {{Relational Non-Arbitrariness}}}.
\newblock \bibinfo{journal}{\emph{Columbia Law Review}} \bibinfo{volume}{119},
  \bibinfo{number}{7} (\bibinfo{date}{Nov.} \bibinfo{year}{2019}),
  \bibinfo{pages}{1773--1792}.
\newblock


\bibitem[Deeks(2019)]%
        {deeks2019judicial}
\bibfield{author}{\bibinfo{person}{Ashley Deeks}.}
  \bibinfo{year}{2019}\natexlab{}.
\newblock \showarticletitle{The {{Judicial Demand}} for {{Explainable
  Artificial Intelligence Essays}}}.
\newblock \bibinfo{journal}{\emph{Columbia Law Review}} \bibinfo{volume}{119},
  \bibinfo{number}{7} (\bibinfo{year}{2019}), \bibinfo{pages}{1829--1850}.
\newblock
\urldef\tempurl%
\url{https://heinonline.org/HOL/P?h=hein.journals/clr119&i=1907}
\showURL{%
\tempurl}


\bibitem[Deloitte(2021)]%
        {deloitte2021study}
\bibfield{author}{\bibinfo{person}{Deloitte}.} \bibinfo{year}{2021}\natexlab{}.
\newblock \bibinfo{booktitle}{\emph{Study to {{Support}} the {{Commission}}'s
  {{Impact Assessment}} on {{Liability}} for {{Artificial Intelligence}}}}.
\newblock \bibinfo{type}{{T}echnical {R}eport}. \bibinfo{institution}{{European
  Commission}}.
\newblock
\urldef\tempurl%
\url{https://doi.org/10.2838/433}
\showDOI{\tempurl}


\bibitem[{Deloitte Center for Regulatory Strategy}(2018)]%
        {deloittecenterforregulatorystrategy2018model}
\bibfield{author}{\bibinfo{person}{{Deloitte Center for Regulatory Strategy}}.}
  \bibinfo{year}{2018}\natexlab{}.
\newblock \bibinfo{booktitle}{\emph{Model Risk Management: {{Building}}
  Supervisory Confidence}}.
\newblock \bibinfo{type}{{T}echnical {R}eport}. \bibinfo{institution}{{Deloitte
  LLP}}.
\newblock
\urldef\tempurl%
\url{https://www2.deloitte.com/content/dam/Deloitte/lu/Documents/risk/lu-model-risk-management.pdf}
\showURL{%
\tempurl}


\bibitem[Desmarais and Lowder(2019)]%
        {desmarais2019pretrial}
\bibfield{author}{\bibinfo{person}{Sarah~L. Desmarais} {and}
  \bibinfo{person}{Evan~M. Lowder}.} \bibinfo{year}{2019}\natexlab{}.
\newblock \bibinfo{booktitle}{\emph{Pretrial {{Risk Assessment Tools}}: {{A
  Primer}} for {{Judges}}, {{Prosecutors}}, and {{Defense Attorneys}}}}.
\newblock \bibinfo{type}{{T}echnical {R}eport}. \bibinfo{institution}{{Safety
  and Justice Challenge}}.
\newblock


\bibitem[{Doshi-Velez} and Kim(2017)]%
        {doshi-velez2017rigorous}
\bibfield{author}{\bibinfo{person}{Finale {Doshi-Velez}} {and}
  \bibinfo{person}{Been Kim}.} \bibinfo{year}{2017}\natexlab{}.
\newblock \showarticletitle{Towards {{A Rigorous Science}} of {{Interpretable
  Machine Learning}}}.
\newblock  (\bibinfo{year}{2017}).
\newblock
\urldef\tempurl%
\url{https://doi.org/10.48550/arxiv.1702.08608}
\showDOI{\tempurl}


\bibitem[{Doshi-Velez} et~al\mbox{.}(2017)]%
        {doshi-velez2017accountability}
\bibfield{author}{\bibinfo{person}{Finale {Doshi-Velez}},
  \bibinfo{person}{Mason Kortz}, \bibinfo{person}{Ryan Budish},
  \bibinfo{person}{Christopher Bavitz}, \bibinfo{person}{Samuel~J. Gershman},
  \bibinfo{person}{David O'Brien}, \bibinfo{person}{Stuart Shieber},
  \bibinfo{person}{Jim Waldo}, \bibinfo{person}{David Weinberger}, {and}
  \bibinfo{person}{Alexandra Wood}.} \bibinfo{year}{2017}\natexlab{}.
\newblock \showarticletitle{Accountability of {{AI Under}} the {{Law}}: {{The
  Role}} of {{Explanation}}}.
\newblock \bibinfo{journal}{\emph{SSRN Electronic Journal}}
  (\bibinfo{year}{2017}).
\newblock
\showISSN{1556-5068}
\urldef\tempurl%
\url{https://doi.org/10.2139/ssrn.3064761}
\showDOI{\tempurl}


\bibitem[Eckhouse et~al\mbox{.}(2019)]%
        {eckhouse2019layers}
\bibfield{author}{\bibinfo{person}{Laurel Eckhouse}, \bibinfo{person}{Kristian
  Lum}, \bibinfo{person}{Cynthia {Conti-Cook}}, {and} \bibinfo{person}{Julie
  Ciccolini}.} \bibinfo{year}{2019}\natexlab{}.
\newblock \showarticletitle{Layers of {{Bias}}: {{A Unified Approach}} for
  {{Understanding Problems With Risk Assessment}}}.
\newblock \bibinfo{journal}{\emph{Criminal Justice and Behavior}}
  \bibinfo{volume}{46}, \bibinfo{number}{2} (\bibinfo{date}{Feb.}
  \bibinfo{year}{2019}), \bibinfo{pages}{185--209}.
\newblock
\showISSN{0093-8548, 1552-3594}
\urldef\tempurl%
\url{https://doi.org/10.1177/0093854818811379}
\showDOI{\tempurl}


\bibitem[Elshawi et~al\mbox{.}(2019)]%
        {elshawi2019interpretability}
\bibfield{author}{\bibinfo{person}{Radwa Elshawi}, \bibinfo{person}{Mouaz~H.
  {Al-Mallah}}, {and} \bibinfo{person}{Sherif Sakr}.}
  \bibinfo{year}{2019}\natexlab{}.
\newblock \showarticletitle{On the Interpretability of Machine Learning-Based
  Model for Predicting Hypertension}.
\newblock \bibinfo{journal}{\emph{BMC Medical Informatics and Decision Making}}
  \bibinfo{volume}{19}, \bibinfo{number}{1} (\bibinfo{date}{Dec.}
  \bibinfo{year}{2019}), \bibinfo{pages}{146}.
\newblock
\showISSN{1472-6947}
\urldef\tempurl%
\url{https://doi.org/10.1186/s12911-019-0874-0}
\showDOI{\tempurl}


\bibitem[{European Commission, Directorate-General for Communications Networks,
  Content {and} Technology}(2020)]%
  {europeancommissiondirectorate-generalforcommunicationsnetworkscontentandtechnology2020european}
\bibfield{author}{\bibinfo{person}{{European Commission, Directorate-General
  for Communications Networks, Content {and} Technology}}.}
  \bibinfo{year}{2020}\natexlab{}.
\newblock \bibinfo{booktitle}{\emph{European Enterprise Survey on the Use of
  Technologies Based on Artificial Intelligence}}.
\newblock \bibinfo{type}{{T}echnical {R}eport}.
\newblock
\urldef\tempurl%
\url{https://data.europa.eu/doi/10.2759/759368}
\showURL{%
\tempurl}


\bibitem[{European Commission Expert Group on Liability {and} New Technologies
  - New Technologies Formation}(2019)]%
        {ecexpertgroup2019liability}
\bibfield{author}{\bibinfo{person}{{European Commission Expert Group on
  Liability {and} New Technologies - New Technologies Formation}}.}
  \bibinfo{year}{2019}\natexlab{}.
\newblock \bibinfo{booktitle}{\emph{Liability for {{Artificial Intelligence}}
  and Other Emerging Digital Technologies}}.
\newblock \bibinfo{type}{{T}echnical {R}eport}.
\newblock


\bibitem[Ficklin et~al\mbox{.}(2020)]%
        {ficklin2020innovation}
\bibfield{author}{\bibinfo{person}{Patrice~Alexander Ficklin},
  \bibinfo{person}{Tom Pahl}, {and} \bibinfo{person}{Paul Watkins}.}
  \bibinfo{year}{2020}\natexlab{}.
\newblock \bibinfo{title}{Innovation Spotlight: {{Providing}} Adverse Action
  Notices When Using {{AI}}/{{ML}} Models}.
\newblock
\newblock
\urldef\tempurl%
\url{https://www.consumerfinance.gov/about-us/blog/innovation-spotlight-providing-adverse-action-notices-when-using-ai-ml-models/}
\showURL{%
\tempurl}


\bibitem[Ficklin and Watkins(2019)]%
        {ficklin2019update}
\bibfield{author}{\bibinfo{person}{Patrice~Alexander Ficklin} {and}
  \bibinfo{person}{Paul Watkins}.} \bibinfo{year}{2019}\natexlab{}.
\newblock \bibinfo{title}{An Update on Credit Access and the {{Bureau}}'s First
  {{No-Action Letter}}}.
\newblock
\newblock
\urldef\tempurl%
\url{https://www.consumerfinance.gov/about-us/blog/update-credit-access-and-no-action-letter/}
\showURL{%
\tempurl}


\bibitem[Flitter(2020)]%
        {flitter2020black}
\bibfield{author}{\bibinfo{person}{Emily Flitter}.}
  \bibinfo{year}{2020}\natexlab{}.
\newblock \showarticletitle{Black {{Homeowners Struggle}} to {{Get Insurers}}
  to {{Pay Claims}}}.
\newblock \bibinfo{journal}{\emph{The New York Times}} (\bibinfo{date}{Dec.}
  \bibinfo{year}{2020}).
\newblock
\showISSN{0362-4331}
\urldef\tempurl%
\url{https://www.nytimes.com/2020/12/29/business/black-homeowners-insurance-claim.html}
\showURL{%
\tempurl}


\bibitem[Frees and Huang(2021)]%
        {frees2021discriminating}
\bibfield{author}{\bibinfo{person}{Edward W.~(Jed) Frees} {and}
  \bibinfo{person}{Fei Huang}.} \bibinfo{year}{2021}\natexlab{}.
\newblock \showarticletitle{The {{Discriminating}} ({{Pricing}}) {{Actuary}}}.
\newblock \bibinfo{journal}{\emph{North American Actuarial Journal}}
  (\bibinfo{date}{Aug.} \bibinfo{year}{2021}), \bibinfo{pages}{1--23}.
\newblock
\showISSN{1092-0277, 2325-0453}
\urldef\tempurl%
\url{https://doi.org/10.1080/10920277.2021.1951296}
\showDOI{\tempurl}


\bibitem[Gebru et~al\mbox{.}(2021)]%
        {gebru2021datasheets}
\bibfield{author}{\bibinfo{person}{Timnit Gebru}, \bibinfo{person}{Jamie
  Morgenstern}, \bibinfo{person}{Briana Vecchione},
  \bibinfo{person}{Jennifer~Wortman Vaughan}, \bibinfo{person}{Hanna Wallach},
  \bibinfo{person}{Hal~Daum{\'e} Iii}, {and} \bibinfo{person}{Kate Crawford}.}
  \bibinfo{year}{2021}\natexlab{}.
\newblock \showarticletitle{Datasheets for Datasets}.
\newblock \bibinfo{journal}{\emph{Commun. ACM}} \bibinfo{volume}{64},
  \bibinfo{number}{12} (\bibinfo{date}{Dec.} \bibinfo{year}{2021}),
  \bibinfo{pages}{86--92}.
\newblock
\showISSN{0001-0782, 1557-7317}
\urldef\tempurl%
\url{https://doi.org/10.1145/3458723}
\showDOI{\tempurl}


\bibitem[Gesser et~al\mbox{.}(2022)]%
        {gesser2022eu}
\bibfield{author}{\bibinfo{person}{Avi Gesser}, \bibinfo{person}{Robert
  Maddox}, \bibinfo{person}{Anna Gressel}, \bibinfo{person}{Frank Colleluori},
  \bibinfo{person}{Tristan Lockwood}, {and} \bibinfo{person}{Michael Pizzi}.}
  \bibinfo{year}{2022}\natexlab{}.
\newblock \bibinfo{title}{The {{EU AI Liability Directive Will Change
  Artificial Intelligence Legal Risks}}}.
\newblock
\newblock
\urldef\tempurl%
\url{https://www.debevoisedatablog.com/2022/10/24/eu-ai-liability-directive/}
\showURL{%
\tempurl}


\bibitem[Gilpin et~al\mbox{.}(2019)]%
        {gilpin2019explaining}
\bibfield{author}{\bibinfo{person}{Leilani~H. Gilpin}, \bibinfo{person}{David
  Bau}, \bibinfo{person}{Ben~Z. Yuan}, \bibinfo{person}{Ayesha Bajwa},
  \bibinfo{person}{Michael Specter}, {and} \bibinfo{person}{Lalana Kagal}.}
  \bibinfo{year}{2019}\natexlab{}.
\newblock \bibinfo{title}{Explaining {{Explanations}}: {{An Overview}} of
  {{Interpretability}} of {{Machine Learning}}}.
\newblock
\newblock
\showeprint[arxiv]{1806.00069}~[cs, stat]
\urldef\tempurl%
\url{http://arxiv.org/abs/1806.00069}
\showURL{%
\tempurl}


\bibitem[Gutierrez and Marchant(2021)]%
        {gutierrez2021global}
\bibfield{author}{\bibinfo{person}{Carlos~Ignacio Gutierrez} {and}
  \bibinfo{person}{Gary Marchant}.} \bibinfo{year}{2021}\natexlab{}.
\newblock \bibinfo{booktitle}{\emph{A {{Global Perspective}} of {{Soft Law
  Programs}} for the {{Governance}} of {{Artificial Intelligence}}}}.
\newblock \bibinfo{type}{{T}echnical {R}eport}. \bibinfo{institution}{{Center
  for Law, Science and Innovation, Sandra Day O'Connor College of Law, Arizona
  State University}}.
\newblock
\urldef\tempurl%
\url{https://lsi.asulaw.org/softlaw/wp-content/uploads/sites/7/2022/08/final-database-report-002-compressed.pdf}
\showURL{%
\tempurl}


\bibitem[Hall et~al\mbox{.}(2021)]%
        {hall2021united}
\bibfield{author}{\bibinfo{person}{Patrick Hall}, \bibinfo{person}{Benjamin
  Cox}, \bibinfo{person}{Steven Dickerson}, \bibinfo{person}{Arjun
  Ravi~Kannan}, \bibinfo{person}{Raghu Kulkarni}, {and}
  \bibinfo{person}{Nicholas Schmidt}.} \bibinfo{year}{2021}\natexlab{}.
\newblock \showarticletitle{A {{United States Fair Lending Perspective}} on
  {{Machine Learning}}}.
\newblock \bibinfo{journal}{\emph{Frontiers in Artificial Intelligence}}
  \bibinfo{volume}{4} (\bibinfo{date}{June} \bibinfo{year}{2021}),
  \bibinfo{pages}{695301}.
\newblock
\showISSN{2624-8212}
\urldef\tempurl%
\url{https://doi.org/10.3389/frai.2021.695301}
\showDOI{\tempurl}


\bibitem[Hall and Gill(2019)]%
        {hall2019introduction}
\bibfield{author}{\bibinfo{person}{Patrick Hall} {and} \bibinfo{person}{Navdeep
  Gill}.} \bibinfo{year}{2019}\natexlab{}.
\newblock \bibinfo{booktitle}{\emph{An {{Introduction}} to {{Machine Learning
  Interpretability}}} (\bibinfo{edition}{second} ed.)}.
\newblock \bibinfo{publisher}{{O'Reilly Media, Inc.}}
\newblock


\bibitem[Hoffman et~al\mbox{.}(2021)]%
        {hoffman2021stakeholder}
\bibfield{author}{\bibinfo{person}{R.R. Hoffman}, \bibinfo{person}{G. Klein},
  \bibinfo{person}{S.T. Mueller}, \bibinfo{person}{M. Jalaeian}, {and}
  \bibinfo{person}{C. Tate}.} \bibinfo{year}{2021}\natexlab{}.
\newblock \bibinfo{booktitle}{\emph{The {{Stakeholder Playbook}} for
  {{Explaining AI Systems}}}}.
\newblock \bibinfo{type}{{T}echnical {R}eport}. \bibinfo{institution}{{DARPA
  Explainable AI Program}}.
\newblock


\bibitem[Hong et~al\mbox{.}(2020)]%
        {hong2020human}
\bibfield{author}{\bibinfo{person}{Sungsoo~Ray Hong}, \bibinfo{person}{Jessica
  Hullman}, {and} \bibinfo{person}{Enrico Bertini}.}
  \bibinfo{year}{2020}\natexlab{}.
\newblock \showarticletitle{Human {{Factors}} in {{Model Interpretability}}:
  {{Industry Practices}}, {{Challenges}}, and {{Needs}}}.
\newblock \bibinfo{journal}{\emph{Proceedings of the ACM on Human-Computer
  Interaction}} \bibinfo{volume}{4}, \bibinfo{number}{CSCW1}
  (\bibinfo{date}{May} \bibinfo{year}{2020}), \bibinfo{pages}{1--26}.
\newblock
\showISSN{2573-0142}
\urldef\tempurl%
\url{https://doi.org/10.1145/3392878}
\showDOI{\tempurl}


\bibitem[Huq(2018)]%
        {huq2018what}
\bibfield{author}{\bibinfo{person}{Aziz~Z. Huq}.}
  \bibinfo{year}{2018}\natexlab{}.
\newblock \showarticletitle{What Is {{Discriminatory Intent}}?}
\newblock \bibinfo{journal}{\emph{Cornell Law Review}}  \bibinfo{volume}{103}
  (\bibinfo{year}{2018}), \bibinfo{pages}{1211--1292}.
\newblock


\bibitem[Jobin et~al\mbox{.}(2019)]%
        {jobin2019global}
\bibfield{author}{\bibinfo{person}{Anna Jobin}, \bibinfo{person}{Marcello
  Ienca}, {and} \bibinfo{person}{Effy Vayena}.}
  \bibinfo{year}{2019}\natexlab{}.
\newblock \showarticletitle{The Global Landscape of {{AI}} Ethics Guidelines}.
\newblock \bibinfo{journal}{\emph{Nature Machine Intelligence}}
  \bibinfo{volume}{1}, \bibinfo{number}{9} (\bibinfo{date}{Sept.}
  \bibinfo{year}{2019}), \bibinfo{pages}{389--399}.
\newblock
\showISSN{2522-5839}
\urldef\tempurl%
\url{https://doi.org/10.1038/s42256-019-0088-2}
\showDOI{\tempurl}


\bibitem[Jones(2023)]%
        {jones2023ai}
\bibfield{author}{\bibinfo{person}{Kate Jones}.}
  \bibinfo{year}{2023}\natexlab{}.
\newblock \bibinfo{booktitle}{\emph{{{AI}} Governance and Human Rights:
  {{Resetting}} the Relationship}}.
\newblock \bibinfo{type}{{T}echnical {R}eport}. \bibinfo{institution}{{Chatham
  House}}.
\newblock
\urldef\tempurl%
\url{https://www.chathamhouse.org/sites/default/files/2023-01/2023-01-10-AI-governance-human-rights-jones.pdf}
\showURL{%
\tempurl}


\bibitem[Karner et~al\mbox{.}(2021)]%
        {karner2021comparative}
\bibfield{author}{\bibinfo{person}{Ernst Karner}, \bibinfo{person}{Bernhard~A.
  Koch}, {and} \bibinfo{person}{Mark~A. Geistfeld}.}
  \bibinfo{year}{2021}\natexlab{}.
\newblock \bibinfo{booktitle}{\emph{Comparative Law Study on Civil Liability
  for Artificial Intelligence}}.
\newblock \bibinfo{type}{{T}echnical {R}eport}. \bibinfo{institution}{{European
  Commission Directorate General for Justice and Consumers.}}
\newblock
\urldef\tempurl%
\url{https://data.europa.eu/doi/10.2838/77360}
\showURL{%
\tempurl}


\bibitem[Kaur et~al\mbox{.}(2020)]%
        {kaur2020interpreting}
\bibfield{author}{\bibinfo{person}{Harmanpreet Kaur}, \bibinfo{person}{Harsha
  Nori}, \bibinfo{person}{Samuel Jenkins}, \bibinfo{person}{Rich Caruana},
  \bibinfo{person}{Hanna Wallach}, {and} \bibinfo{person}{Jennifer
  Wortman~Vaughan}.} \bibinfo{year}{2020}\natexlab{}.
\newblock \showarticletitle{Interpreting {{Interpretability}}: {{Understanding
  Data Scientists}}' {{Use}} of {{Interpretability Tools}} for {{Machine
  Learning}}}. In \bibinfo{booktitle}{\emph{Proceedings of the 2020 {{CHI
  Conference}} on {{Human Factors}} in {{Computing Systems}}}}.
  \bibinfo{publisher}{{ACM}}, \bibinfo{address}{{Honolulu HI USA}},
  \bibinfo{pages}{1--14}.
\newblock
\showISBNx{978-1-4503-6708-0}
\urldef\tempurl%
\url{https://doi.org/10.1145/3313831.3376219}
\showDOI{\tempurl}


\bibitem[Kindermans et~al\mbox{.}(2019)]%
        {kindermans2019reliability}
\bibfield{author}{\bibinfo{person}{Pieter-Jan Kindermans},
  \bibinfo{person}{Sara Hooker}, \bibinfo{person}{Julius Adebayo},
  \bibinfo{person}{Maximilian Alber}, \bibinfo{person}{Kristof~T. Sch{\"u}tt},
  \bibinfo{person}{Sven D{\"a}hne}, \bibinfo{person}{Dumitru Erhan}, {and}
  \bibinfo{person}{Been Kim}.} \bibinfo{year}{2019}\natexlab{}.
\newblock \showarticletitle{The ({{Un}})Reliability of {{Saliency Methods}}}.
\newblock In \bibinfo{booktitle}{\emph{Explainable {{AI}}: {{Interpreting}},
  {{Explaining}} and {{Visualizing Deep Learning}}}},
  \bibfield{editor}{\bibinfo{person}{Wojciech Samek},
  \bibinfo{person}{Gr{\'e}goire Montavon}, \bibinfo{person}{Andrea Vedaldi},
  \bibinfo{person}{Lars~Kai Hansen}, {and} \bibinfo{person}{Klaus-Robert
  M{\"u}ller}} (Eds.). Vol.~\bibinfo{volume}{11700}.
  \bibinfo{publisher}{{Springer International Publishing}},
  \bibinfo{address}{{Cham}}, \bibinfo{pages}{267--280}.
\newblock
\showISBNx{978-3-030-28953-9 978-3-030-28954-6}
\urldef\tempurl%
\url{https://doi.org/10.1007/978-3-030-28954-6_14}
\showDOI{\tempurl}


\bibitem[Kiritz and Sarfati(2018)]%
        {kiritz2018supervisory}
\bibfield{author}{\bibinfo{person}{Nicholas Kiritz} {and}
  \bibinfo{person}{Philippe Sarfati}.} \bibinfo{year}{2018}\natexlab{}.
\newblock \showarticletitle{Supervisory {{Guidance}} on {{Model Risk
  Management}} ({{SR}} 11-7) {{Versus Enterprise-Wide Model Risk Management}}
  for {{Deposit-Taking Institutions}} ({{E-23}}): {{A Detailed Comparative
  Analysis}}}.
\newblock \bibinfo{journal}{\emph{SSRN Electronic Journal}}
  (\bibinfo{year}{2018}).
\newblock
\showISSN{1556-5068}
\urldef\tempurl%
\url{https://doi.org/10.2139/ssrn.3332484}
\showDOI{\tempurl}


\bibitem[Klein and Rudolph(2019)]%
        {klein2019company}
\bibfield{author}{\bibinfo{person}{Al Klein} {and} \bibinfo{person}{Karen
  Rudolph}.} \bibinfo{year}{2019}\natexlab{}.
\newblock \bibinfo{booktitle}{\emph{Company {{Practice Survey}} of {{Individual
  Life Insurance Accelerated Underwriting}} \textendash{} {{Preliminary
  Results}}}}.
\newblock \bibinfo{type}{{T}echnical {R}eport}. \bibinfo{institution}{{Society
  of Actuaries}}.
\newblock
\urldef\tempurl%
\url{https://www.soa.org/globalassets/assets/files/resources/research-report/2019/accelerated-underwriting-preliminary-results.pdf}
\showURL{%
\tempurl}


\bibitem[K{\"o}nig and Wenzelburger(2021)]%
        {konig2021whenb}
\bibfield{author}{\bibinfo{person}{Pascal~D K{\"o}nig} {and}
  \bibinfo{person}{Georg Wenzelburger}.} \bibinfo{year}{2021}\natexlab{}.
\newblock \showarticletitle{When {{Politicization Stops Algorithms}} in
  {{Criminal Justice}}}.
\newblock \bibinfo{journal}{\emph{The British Journal of Criminology}}
  \bibinfo{volume}{61}, \bibinfo{number}{3} (\bibinfo{date}{April}
  \bibinfo{year}{2021}), \bibinfo{pages}{832--851}.
\newblock
\showISSN{0007-0955, 1464-3529}
\urldef\tempurl%
\url{https://doi.org/10.1093/bjc/azaa099}
\showDOI{\tempurl}


\bibitem[Krafcheck(2021)]%
        {krafcheck2021insurance}
\bibfield{author}{\bibinfo{person}{Eric~P. Krafcheck}.}
  \bibinfo{year}{2021}\natexlab{}.
\newblock \bibinfo{title}{The Insurance Industry's Renewed Focus on Disparate
  Impacts and Unfair Discrimination}.
\newblock
\newblock
\urldef\tempurl%
\url{https://nl.milliman.com/nl-NL/insight/the-insurance-industrys-renewed-focus-on-disparate-impacts-and-unfair-discrimination}
\showURL{%
\tempurl}


\bibitem[Krishna et~al\mbox{.}(2022)]%
        {krishna2022disagreement}
\bibfield{author}{\bibinfo{person}{Satyapriya Krishna}, \bibinfo{person}{Tessa
  Han}, \bibinfo{person}{Alex Gu}, \bibinfo{person}{Javin Pombra},
  \bibinfo{person}{Shahin Jabbari}, \bibinfo{person}{Steven Wu}, {and}
  \bibinfo{person}{Himabindu Lakkaraju}.} \bibinfo{year}{2022}\natexlab{}.
\newblock \bibinfo{title}{The {{Disagreement Problem}} in {{Explainable Machine
  Learning}}: {{A Practitioner}}'s {{Perspective}}}.
\newblock
\newblock
\urldef\tempurl%
\url{https://doi.org/10.48550/arXiv.2202.01602}
\showDOI{\tempurl}
\showeprint[arxiv]{2202.01602}~[cs]


\bibitem[Krivorotov and Richey(2022)]%
        {krivorotov2022explaining}
\bibfield{author}{\bibinfo{person}{George Krivorotov} {and}
  \bibinfo{person}{Jeremiah Richey}.} \bibinfo{year}{2022}\natexlab{}.
\newblock \showarticletitle{Explaining {{Denials}}: {{Adverse Action Codes}}
  and {{Machine Learning}} in {{Credit Decisioning}}}.
\newblock \bibinfo{journal}{\emph{SSRN Electronic Journal}}
  (\bibinfo{year}{2022}).
\newblock
\showISSN{1556-5068}
\urldef\tempurl%
\url{https://doi.org/10.2139/ssrn.4133915}
\showDOI{\tempurl}


\bibitem[Lakkaraju and Bastani(2020)]%
        {lakkaraju2020how}
\bibfield{author}{\bibinfo{person}{Himabindu Lakkaraju} {and}
  \bibinfo{person}{Osbert Bastani}.} \bibinfo{year}{2020}\natexlab{}.
\newblock \showarticletitle{"{{How}} Do {{I}} Fool You?": {{Manipulating User
  Trust}} via {{Misleading Black Box Explanations}}}. In
  \bibinfo{booktitle}{\emph{Proceedings of the {{AAAI}}/{{ACM Conference}} on
  {{AI}}, {{Ethics}}, and {{Society}}}}. \bibinfo{publisher}{{ACM}},
  \bibinfo{address}{{New York NY USA}}, \bibinfo{pages}{79--85}.
\newblock
\showISBNx{978-1-4503-7110-0}
\urldef\tempurl%
\url{https://doi.org/10.1145/3375627.3375833}
\showDOI{\tempurl}


\bibitem[Legislature(2019a)]%
        {idaholegislature2019house}
\bibfield{author}{\bibinfo{person}{Idaho Legislature}.}
  \bibinfo{year}{2019}\natexlab{a}.
\newblock \bibinfo{title}{House {{Bill}} 118 (2019 {{Legislation}})}.
\newblock
\newblock
\urldef\tempurl%
\url{https://legislature.idaho.gov/sessioninfo/2019/legislation/H0118/}
\showURL{%
\tempurl}


\bibitem[Legislature(2019b)]%
        {idaholegislature2019proceedings}
\bibfield{author}{\bibinfo{person}{Idaho Legislature}.}
  \bibinfo{year}{2019}\natexlab{b}.
\newblock \bibinfo{title}{Proceedings of the {{House Judiciary}}, {{Rules}} \&
  {{Administration Committee}} - {{February}} 19, 2019}.
\newblock
\newblock
\urldef\tempurl%
\url{https://legislature.idaho.gov/sessioninfo/2019/standingcommittees/HJUD/}
\showURL{%
\tempurl}


\bibitem[Legislature(2019c)]%
        {idaholegislature2019proceedingsa}
\bibfield{author}{\bibinfo{person}{Idaho Legislature}.}
  \bibinfo{year}{2019}\natexlab{c}.
\newblock \bibinfo{title}{Proceedings of the {{House Judiciary}}, {{Rules}} \&
  {{Administration Committee}} - {{March}} 19, 2019}.
\newblock
\newblock
\urldef\tempurl%
\url{https://legislature.idaho.gov/sessioninfo/2019/standingcommittees/HJUD/}
\showURL{%
\tempurl}


\bibitem[Legislature(2019d)]%
        {idaholegislature2019proceedingsb}
\bibfield{author}{\bibinfo{person}{Idaho Legislature}.}
  \bibinfo{year}{2019}\natexlab{d}.
\newblock \bibinfo{title}{Proceedings of the {{Senate Judiciary}} \& {{Rules
  Committee}} - {{March}} 11, 2019}.
\newblock
\newblock
\urldef\tempurl%
\url{https://legislature.idaho.gov/sessioninfo/2019/standingcommittees/SJR/}
\showURL{%
\tempurl}


\bibitem[Legislature(2022a)]%
        {oklahomastatelegislature2022bill}
\bibfield{author}{\bibinfo{person}{Oklahoma~State Legislature}.}
  \bibinfo{year}{2022}\natexlab{a}.
\newblock \bibinfo{title}{Bill {{Information}} for {{HB}} 3186}.
\newblock
\newblock
\urldef\tempurl%
\url{http://www.oklegislature.gov/BillInfo.aspx?Bill=hb3186&Session=2200}
\showURL{%
\tempurl}


\bibitem[Legislature(2022b)]%
        {rhodeislandlegislature2022legislative}
\bibfield{author}{\bibinfo{person}{Rhode~Island Legislature}.}
  \bibinfo{year}{2022}\natexlab{b}.
\newblock \bibinfo{title}{Legislative {{Status Report}}: {{House Bill No}}.
  7230}.
\newblock
\newblock
\urldef\tempurl%
\url{https://status.rilegislature.gov/}
\showURL{%
\tempurl}


\bibitem[Liao et~al\mbox{.}(2020)]%
        {liao2020questioning}
\bibfield{author}{\bibinfo{person}{Q.~Vera Liao}, \bibinfo{person}{Daniel
  Gruen}, {and} \bibinfo{person}{Sarah Miller}.}
  \bibinfo{year}{2020}\natexlab{}.
\newblock \showarticletitle{Questioning the {{AI}}: {{Informing Design
  Practices}} for {{Explainable AI User Experiences}}}. In
  \bibinfo{booktitle}{\emph{Proceedings of the 2020 {{CHI Conference}} on
  {{Human Factors}} in {{Computing Systems}}}}. \bibinfo{publisher}{{ACM}},
  \bibinfo{address}{{Honolulu HI USA}}, \bibinfo{pages}{1--15}.
\newblock
\showISBNx{978-1-4503-6708-0}
\urldef\tempurl%
\url{https://doi.org/10.1145/3313831.3376590}
\showDOI{\tempurl}


\bibitem[Lipton(2019)]%
        {lipton2019idahoa}
\bibfield{author}{\bibinfo{person}{Beryl Lipton}.}
  \bibinfo{year}{2019}\natexlab{}.
\newblock \showarticletitle{Idaho Legislators Approve Law Requiring
  Transparency for Risk Assessment Tools}.
\newblock \bibinfo{journal}{\emph{MuckRock}} (\bibinfo{date}{March}
  \bibinfo{year}{2019}).
\newblock
\urldef\tempurl%
\url{https://www.muckrock.com/news/archives/2019/mar/26/algorithms-idaho-bill-update/}
\showURL{%
\tempurl}


\bibitem[Lipton(2018)]%
        {lipton2018mythos}
\bibfield{author}{\bibinfo{person}{Zachary~C. Lipton}.}
  \bibinfo{year}{2018}\natexlab{}.
\newblock \showarticletitle{The {{Mythos}} of {{Model Interpretability}}:
  {{In}} Machine Learning, the Concept of Interpretability Is Both Important
  and Slippery.}
\newblock \bibinfo{journal}{\emph{Queue}} \bibinfo{volume}{16},
  \bibinfo{number}{3} (\bibinfo{date}{June} \bibinfo{year}{2018}),
  \bibinfo{pages}{31--57}.
\newblock
\showISSN{1542-7730, 1542-7749}
\urldef\tempurl%
\url{https://doi.org/10.1145/3236386.3241340}
\showDOI{\tempurl}


\bibitem[Lombrozo(2006)]%
        {lombrozo2006structure}
\bibfield{author}{\bibinfo{person}{Tania Lombrozo}.}
  \bibinfo{year}{2006}\natexlab{}.
\newblock \showarticletitle{The Structure and Function of Explanations}.
\newblock \bibinfo{journal}{\emph{Trends in Cognitive Sciences}}
  \bibinfo{volume}{10}, \bibinfo{number}{10} (\bibinfo{date}{Oct.}
  \bibinfo{year}{2006}), \bibinfo{pages}{464--470}.
\newblock
\showISSN{13646613}
\urldef\tempurl%
\url{https://doi.org/10.1016/j.tics.2006.08.004}
\showDOI{\tempurl}


\bibitem[Mac et~al\mbox{.}(2021)]%
        {mac2021your}
\bibfield{author}{\bibinfo{person}{Ryan Mac}, \bibinfo{person}{Caroline
  Haskins}, \bibinfo{person}{Brianna Sacks}, {and} \bibinfo{person}{Logan
  McDonald}.} \bibinfo{year}{2021}\natexlab{}.
\newblock \bibinfo{title}{Your {{Local Police Department Might Have Used This
  Facial Recognition Tool To Surveil You}}. {{Find Out Here}}.}
\newblock
\newblock
\urldef\tempurl%
\url{https://www.buzzfeednews.com/article/ryanmac/facial-recognition-local-police-clearview-ai-table}
\showURL{%
\tempurl}


\bibitem[Marchant(2011)]%
        {marchant2011addressing}
\bibfield{author}{\bibinfo{person}{Gary~E. Marchant}.}
  \bibinfo{year}{2011}\natexlab{}.
\newblock \showarticletitle{Addressing the {{Pacing Problem}}}.
\newblock In \bibinfo{booktitle}{\emph{The {{Growing Gap Between Emerging
  Technologies}} and {{Legal-Ethical Oversight}}}},
  \bibfield{editor}{\bibinfo{person}{Gary~E. Marchant},
  \bibinfo{person}{Braden~R. Allenby}, {and} \bibinfo{person}{Joseph~R.
  Herkert}} (Eds.). Vol.~\bibinfo{volume}{7}. \bibinfo{publisher}{{Springer
  Netherlands}}, \bibinfo{address}{{Dordrecht}}, \bibinfo{pages}{199--205}.
\newblock
\showISBNx{978-94-007-1355-0 978-94-007-1356-7}
\urldef\tempurl%
\url{https://doi.org/10.1007/978-94-007-1356-7_13}
\showDOI{\tempurl}


\bibitem[Marchant and Tournas(2019)]%
        {marchant2019ai}
\bibfield{author}{\bibinfo{person}{Gary~E. Marchant} {and}
  \bibinfo{person}{Lucille~M. Tournas}.} \bibinfo{year}{2019}\natexlab{}.
\newblock \showarticletitle{{{AI Health Care Liability}}: {{From Research
  Trials}} to {{Court Trials}}}.
\newblock \bibinfo{journal}{\emph{Journal of Health \& Life Sciences Law}}
  \bibinfo{volume}{12}, \bibinfo{number}{2} (\bibinfo{date}{Feb.}
  \bibinfo{year}{2019}), \bibinfo{pages}{23--42}.
\newblock


\bibitem[McElfresh and Yazdi(2022)]%
        {mcelfresh2022machine}
\bibfield{author}{\bibinfo{person}{Duncan McElfresh} {and}
  \bibinfo{person}{Sormeh Yazdi}.} \bibinfo{year}{2022}\natexlab{}.
\newblock \bibinfo{booktitle}{\emph{Machine {{Learning Explainability}} \&
  {{Fairness}}: {{Insights}} from {{Consumer Lending}}}}.
\newblock \bibinfo{type}{{T}echnical {R}eport}.
  \bibinfo{institution}{{FinRegLab}}.
\newblock
\urldef\tempurl%
\url{https://finreglab.org/wp-content/uploads/2022/04/FinRegLab_Stanford_ML-Explainability-and-Fairness_Insights-from-Consumer-Lending-April-2022.pdf}
\showURL{%
\tempurl}


\bibitem[Medine(2020)]%
        {medine2020there}
\bibfield{author}{\bibinfo{person}{Taylor Medine}.}
  \bibinfo{year}{2020}\natexlab{}.
\newblock \bibinfo{title}{Is There a Life Insurance Race Gap?}
\newblock
\newblock
\urldef\tempurl%
\url{https://havenlife.com/blog/life-insurance-racial-wealth-gap-statistics/}
\showURL{%
\tempurl}


\bibitem[Metaxa and Hancock(2022)]%
        {metaxa2022using}
\bibfield{author}{\bibinfo{person}{Dana{\"e} Metaxa} {and}
  \bibinfo{person}{Jeff Hancock}.} \bibinfo{year}{2022}\natexlab{}.
\newblock \bibinfo{booktitle}{\emph{Using {{Algorithm Audits}} to {{Understand
  AI}}}}.
\newblock \bibinfo{type}{Policy {{Brief}}}. \bibinfo{institution}{{Stanford
  University Human-Centered Artificial Intelligence}}.
\newblock
\urldef\tempurl%
\url{https://hai.stanford.edu/policy-brief-using-algorithm-audits-understand-ai}
\showURL{%
\tempurl}


\bibitem[Metcalf et~al\mbox{.}(2021)]%
        {metcalf2021algorithmic}
\bibfield{author}{\bibinfo{person}{Jacob Metcalf}, \bibinfo{person}{Emanuel
  Moss}, \bibinfo{person}{Elizabeth~Anne Watkins}, \bibinfo{person}{Ranjit
  Singh}, {and} \bibinfo{person}{Madeleine~Clare Elish}.}
  \bibinfo{year}{2021}\natexlab{}.
\newblock \showarticletitle{Algorithmic {{Impact Assessments}} and
  {{Accountability}}: {{The Co-construction}} of {{Impacts}}}. In
  \bibinfo{booktitle}{\emph{Proceedings of the 2021 {{ACM Conference}} on
  {{Fairness}}, {{Accountability}}, and {{Transparency}}}}.
  \bibinfo{publisher}{{ACM}}, \bibinfo{address}{{Virtual Event Canada}},
  \bibinfo{pages}{735--746}.
\newblock
\showISBNx{978-1-4503-8309-7}
\urldef\tempurl%
\url{https://doi.org/10.1145/3442188.3445935}
\showDOI{\tempurl}


\bibitem[Miller(2019)]%
        {miller2019explanation}
\bibfield{author}{\bibinfo{person}{Tim Miller}.}
  \bibinfo{year}{2019}\natexlab{}.
\newblock \showarticletitle{Explanation in Artificial Intelligence:
  {{Insights}} from the Social Sciences}.
\newblock \bibinfo{journal}{\emph{Artificial Intelligence}}
  \bibinfo{volume}{267} (\bibinfo{date}{Feb.} \bibinfo{year}{2019}),
  \bibinfo{pages}{1--38}.
\newblock
\showISSN{00043702}
\urldef\tempurl%
\url{https://doi.org/10.1016/j.artint.2018.07.007}
\showDOI{\tempurl}


\bibitem[Miller et~al\mbox{.}(2017)]%
        {miller2017explainable}
\bibfield{author}{\bibinfo{person}{Tim Miller}, \bibinfo{person}{Piers Howe},
  {and} \bibinfo{person}{Liz Sonenberg}.} \bibinfo{year}{2017}\natexlab{}.
\newblock \showarticletitle{Explainable {{AI}}: {{Beware}} of {{Inmates
  Running}} the {{Asylum Or}}: {{How I Learnt}} to {{Stop Worrying}} and
  {{Love}} the {{Social}} and {{Behavioural Sciences}}}.
\newblock  (\bibinfo{year}{2017}).
\newblock
\urldef\tempurl%
\url{https://doi.org/10.48550/ARXIV.1712.00547}
\showDOI{\tempurl}


\bibitem[Mitchell et~al\mbox{.}(2019)]%
        {mitchell2019model}
\bibfield{author}{\bibinfo{person}{Margaret Mitchell}, \bibinfo{person}{Simone
  Wu}, \bibinfo{person}{Andrew Zaldivar}, \bibinfo{person}{Parker Barnes},
  \bibinfo{person}{Lucy Vasserman}, \bibinfo{person}{Ben Hutchinson},
  \bibinfo{person}{Elena Spitzer}, \bibinfo{person}{Inioluwa~Deborah Raji},
  {and} \bibinfo{person}{Timnit Gebru}.} \bibinfo{year}{2019}\natexlab{}.
\newblock \showarticletitle{Model {{Cards}} for {{Model Reporting}}}. In
  \bibinfo{booktitle}{\emph{Proceedings of the {{Conference}} on {{Fairness}},
  {{Accountability}}, and {{Transparency}}}}. \bibinfo{publisher}{{ACM}},
  \bibinfo{address}{{Atlanta GA USA}}, \bibinfo{pages}{220--229}.
\newblock
\showISBNx{978-1-4503-6125-5}
\urldef\tempurl%
\url{https://doi.org/10.1145/3287560.3287596}
\showDOI{\tempurl}


\bibitem[Moss et~al\mbox{.}(2021)]%
        {moss2021assembling}
\bibfield{author}{\bibinfo{person}{Emanuel Moss},
  \bibinfo{person}{Elizabeth~Anne Watkins}, \bibinfo{person}{Ranjit Singh},
  \bibinfo{person}{Madeleine~Clare Elish}, {and} \bibinfo{person}{Jacob
  Metcalf}.} \bibinfo{year}{2021}\natexlab{}.
\newblock \bibinfo{booktitle}{\emph{Assembling {{Accountability}}:
  {{Algorithmic Impact Assessment}} for the {{Public Interest}}}}.
\newblock \bibinfo{type}{{T}echnical {R}eport}. \bibinfo{institution}{{Data \&
  Society}}.
\newblock
\urldef\tempurl%
\url{https://datasociety.net/wp-content/uploads/2021/06/Assembling-Accountability.pdf}
\showURL{%
\tempurl}


\bibitem[Mueller et~al\mbox{.}(2021)]%
        {mueller2021principles}
\bibfield{author}{\bibinfo{person}{Shane~T. Mueller},
  \bibinfo{person}{Elizabeth~S. Veinott}, \bibinfo{person}{Robert~R. Hoffman},
  \bibinfo{person}{Gary Klein}, \bibinfo{person}{Lamia Alam},
  \bibinfo{person}{Tauseef Mamun}, {and} \bibinfo{person}{William~J. Clancey}.}
  \bibinfo{year}{2021}\natexlab{}.
\newblock \showarticletitle{Principles of {{Explanation}} in {{Human-AI
  Systems}}}.
\newblock  (\bibinfo{year}{2021}).
\newblock
\urldef\tempurl%
\url{https://doi.org/10.48550/ARXIV.2102.04972}
\showDOI{\tempurl}


\bibitem[Naiseh et~al\mbox{.}(2020)]%
        {naiseh2020personalising}
\bibfield{author}{\bibinfo{person}{Mohammad Naiseh}, \bibinfo{person}{Nan
  Jiang}, \bibinfo{person}{Jianbing Ma}, {and} \bibinfo{person}{Raian Ali}.}
  \bibinfo{year}{2020}\natexlab{}.
\newblock \showarticletitle{Personalising {{Explainable Recommendations}}:
  {{Literature}} and {{Conceptualisation}}}.
\newblock In \bibinfo{booktitle}{\emph{Trends and {{Innovations}} in
  {{Information Systems}} and {{Technologies}}}},
  \bibfield{editor}{\bibinfo{person}{{\'A}lvaro Rocha}, \bibinfo{person}{Hojjat
  Adeli}, \bibinfo{person}{Lu{\'i}s~Paulo Reis}, \bibinfo{person}{Sandra
  Costanzo}, \bibinfo{person}{Irena Orovic}, {and} \bibinfo{person}{Fernando
  Moreira}} (Eds.). Vol.~\bibinfo{volume}{1160}. \bibinfo{publisher}{{Springer
  International Publishing}}, \bibinfo{address}{{Cham}},
  \bibinfo{pages}{518--533}.
\newblock
\showISBNx{978-3-030-45690-0 978-3-030-45691-7}
\urldef\tempurl%
\url{https://doi.org/10.1007/978-3-030-45691-7_49}
\showDOI{\tempurl}


\bibitem[{National Conference of State Legislatures}(2022a)]%
        {nationalconferenceofstatelegislatures2022legislation}
\bibfield{author}{\bibinfo{person}{{National Conference of State
  Legislatures}}.} \bibinfo{year}{2022}\natexlab{a}.
\newblock \bibinfo{title}{Legislation {{Related}} to {{Artifical
  Intelligence}}}.
\newblock
\newblock
\urldef\tempurl%
\url{https://www.ncsl.org/research/telecommunications-and-information-technology/2020-legislation-related-to-artificial-intelligence.aspx}
\showURL{%
\tempurl}


\bibitem[{National Conference of State Legislatures}(2022b)]%
        {nationalconferenceofstatelegislatures2022racial}
\bibfield{author}{\bibinfo{person}{{National Conference of State
  Legislatures}}.} \bibinfo{year}{2022}\natexlab{b}.
\newblock \bibinfo{booktitle}{\emph{Racial and {{Ethnic Disparities}} in the
  {{Criminal Justice System}}}}.
\newblock \bibinfo{type}{{T}echnical {R}eport}.
\newblock
\urldef\tempurl%
\url{https://www.ncsl.org/civil-and-criminal-justice/racial-and-ethnic-disparities-in-the-criminal-justice-system}
\showURL{%
\tempurl}


\bibitem[Neace(2022)]%
        {neace2022aivia}
\bibfield{author}{\bibinfo{person}{Gabrielle Neace}.}
  \bibinfo{year}{2022}\natexlab{}.
\newblock \bibinfo{title}{{{AIVIA}}: {{A Step Towards Protecting Data Privacy}}
  or a {{Continuation}} of the {{Push}} for {{Individuals}} to {{Trade}} Their
  {{Privacy Rights}} for {{Employment}}?}
\newblock
\newblock
\urldef\tempurl%
\url{https://lawreview.law.uic.edu/news-stories/aivia-a-step-towards-protecting-data-privacy-or-a-continuation-of-the-push-for-individuals-to-trade-their-privacy-rights-for-employment/#}
\showURL{%
\tempurl}


\bibitem[Neely et~al\mbox{.}(2021)]%
        {neely2021order}
\bibfield{author}{\bibinfo{person}{Michael Neely}, \bibinfo{person}{Stefan~F.
  Schouten}, \bibinfo{person}{Maurits J.~R. Bleeker}, {and}
  \bibinfo{person}{Ana Lucic}.} \bibinfo{year}{2021}\natexlab{}.
\newblock \showarticletitle{Order in the {{Court}}: {{Explainable AI Methods
  Prone}} to {{Disagreement}}}. In \bibinfo{booktitle}{\emph{Proc. {{ICML
  Workshop}} on {{Theoretical Foundation}}, {{Criticism}}, and {{Application
  Trends}} of {{Explainable AI}}}}.
\newblock
\showeprint[arxiv]{2105.03287}~[cs]
\urldef\tempurl%
\url{http://arxiv.org/abs/2105.03287}
\showURL{%
\tempurl}


\bibitem[Nonnecke and Carlton(2022)]%
        {nonnecke2022eu}
\bibfield{author}{\bibinfo{person}{Brandie Nonnecke} {and}
  \bibinfo{person}{Camille Carlton}.} \bibinfo{year}{2022}\natexlab{}.
\newblock \showarticletitle{{{EU}} and {{US}} Legislation Seek to Open up
  Digital Platform Data}.
\newblock \bibinfo{journal}{\emph{Science}} \bibinfo{volume}{375},
  \bibinfo{number}{6581} (\bibinfo{date}{Feb.} \bibinfo{year}{2022}),
  \bibinfo{pages}{610--612}.
\newblock
\showISSN{0036-8075, 1095-9203}
\urldef\tempurl%
\url{https://doi.org/10.1126/science.abl8537}
\showDOI{\tempurl}


\bibitem[OECD(2020)]%
        {oecd2020recommendation}
\bibfield{author}{\bibinfo{person}{OECD}.} \bibinfo{year}{2020}\natexlab{}.
\newblock \bibinfo{booktitle}{\emph{Recommendation of the {{Council}} on
  {{Artificial Intelligence}}}}.
\newblock \bibinfo{type}{{T}echnical {R}eport} OECD/LEGAL/0449.
  \bibinfo{institution}{{OECD}}.
\newblock
\urldef\tempurl%
\url{https://legalinstruments.oecd.org/en/instruments/OECD-LEGAL-0449#adherents}
\showURL{%
\tempurl}


\bibitem[Office(2019)]%
        {adacountysheriffsoffice2019ada}
\bibfield{author}{\bibinfo{person}{Ada County~Sheriff's Office}.}
  \bibinfo{year}{2019}\natexlab{}.
\newblock \showarticletitle{Ada {{County}} Is Now Using a Revised Inmate
  {{Pretrial}} Risk Assessment}.
\newblock  (\bibinfo{date}{Nov.} \bibinfo{year}{2019}).
\newblock
\urldef\tempurl%
\url{https://adacounty.id.gov/sheriff/news/ada-county-is-now-using-a-revised-inmate-pretrial-risk-assessment/}
\showURL{%
\tempurl}


\bibitem[Organization(2022)]%
        {internationalstandardsorganization2022iso}
\bibfield{author}{\bibinfo{person}{International~Standards Organization}.}
  \bibinfo{year}{2022}\natexlab{}.
\newblock \bibinfo{title}{{{ISO}}/{{IEC AWI TS}} 6254: {{Information}}
  Technology \textemdash{} {{Artificial}} Intelligence \textemdash{}
  {{Objectives}} and Approaches for Explainability of {{ML}} Models and {{AI}}
  Systems}.
\newblock
\newblock
\urldef\tempurl%
\url{https://www.iso.org/standard/82148.html}
\showURL{%
\tempurl}


\bibitem[O'Shaughnessy et~al\mbox{.}(2022)]%
        {oshaughnessy2022what}
\bibfield{author}{\bibinfo{person}{Matthew O'Shaughnessy},
  \bibinfo{person}{Daniel Schiff}, \bibinfo{person}{Lav Varshney},
  \bibinfo{person}{Christopher Rozell}, {and} \bibinfo{person}{Mark
  Davenport}.} \bibinfo{year}{2022}\natexlab{}.
\newblock \showarticletitle{What Governs Attitudes toward Artificial
  Intelligence Adoption and Governance?}
\newblock \bibinfo{journal}{\emph{Science and Public Policy}}
  (\bibinfo{date}{Oct.} \bibinfo{year}{2022}).
\newblock
\showISSN{0302-3427, 1471-5430}
\urldef\tempurl%
\url{https://doi.org/10.1093/scipol/scac056}
\showDOI{\tempurl}


\bibitem[Pace(2022a)]%
        {pace2022dark}
\bibfield{author}{\bibinfo{person}{Richard Pace}.}
  \bibinfo{year}{2022}\natexlab{a}.
\newblock \bibinfo{title}{Dark {{Skies Ahead}}: {{The CFPB}}'s {{Brewing
  Algorithmic Storm}}}.
\newblock
\newblock
\urldef\tempurl%
\url{https://www.paceanalyticsllc.com/post/cfpb-brewing-algorithmic-storm}
\showURL{%
\tempurl}


\bibitem[Pace(2022b)]%
        {pace2022using}
\bibfield{author}{\bibinfo{person}{Richard Pace}.}
  \bibinfo{year}{2022}\natexlab{b}.
\newblock \bibinfo{title}{Using {{Explainable AI}} to {{Produce ECOA Adverse
  Action Reasons}}: {{What Are The Risks}}?}
\newblock
\newblock
\urldef\tempurl%
\url{https://www.paceanalyticsllc.com/post/ecoa-adverse-actions-and-explainable-ai}
\showURL{%
\tempurl}


\bibitem[Phillips et~al\mbox{.}(2021)]%
        {phillips2021four}
\bibfield{author}{\bibinfo{person}{P.~Jonathon Phillips},
  \bibinfo{person}{Carina~A. Hahn}, \bibinfo{person}{Peter~C. Fontana},
  \bibinfo{person}{Amy~N. Yates}, \bibinfo{person}{Kristen Greene},
  \bibinfo{person}{David~A. Broniatowski}, {and} \bibinfo{person}{Mark~A.
  Przybocki}.} \bibinfo{year}{2021}\natexlab{}.
\newblock \bibinfo{booktitle}{\emph{Four {{Principles}} of {{Explainable
  Artificial Intelligence}}}}.
\newblock \bibinfo{type}{{T}echnical {R}eport}. \bibinfo{institution}{{National
  Institute of Standards and Technology}}.
\newblock
\urldef\tempurl%
\url{https://doi.org/10.6028/NIST.IR.8312}
\showDOI{\tempurl}


\bibitem[Porter et~al\mbox{.}(2022)]%
        {porter2022distinguishing}
\bibfield{author}{\bibinfo{person}{Zoe Porter}, \bibinfo{person}{Annette
  Zimmermann}, \bibinfo{person}{Phillip Morgan}, \bibinfo{person}{John
  McDermid}, \bibinfo{person}{Tom Lawton}, {and} \bibinfo{person}{Ibrahim
  Habli}.} \bibinfo{year}{2022}\natexlab{}.
\newblock \showarticletitle{Distinguishing Two Features of Accountability for
  {{AI}} Technologies}.
\newblock \bibinfo{journal}{\emph{Nature Machine Intelligence}}
  \bibinfo{volume}{4}, \bibinfo{number}{9} (\bibinfo{date}{Sept.}
  \bibinfo{year}{2022}), \bibinfo{pages}{734--736}.
\newblock
\showISSN{2522-5839}
\urldef\tempurl%
\url{https://doi.org/10.1038/s42256-022-00533-0}
\showDOI{\tempurl}


\bibitem[{Poursabzi-Sangdeh} et~al\mbox{.}(2021)]%
        {poursabzi-sangdeh2021manipulating}
\bibfield{author}{\bibinfo{person}{Forough {Poursabzi-Sangdeh}},
  \bibinfo{person}{Daniel~G Goldstein}, \bibinfo{person}{Jake~M Hofman},
  \bibinfo{person}{Jennifer~Wortman Wortman~Vaughan}, {and}
  \bibinfo{person}{Hanna Wallach}.} \bibinfo{year}{2021}\natexlab{}.
\newblock \showarticletitle{Manipulating and {{Measuring Model
  Interpretability}}}. In \bibinfo{booktitle}{\emph{Proceedings of the 2021
  {{CHI Conference}} on {{Human Factors}} in {{Computing Systems}}}}.
  \bibinfo{publisher}{{ACM}}, \bibinfo{address}{{Yokohama Japan}},
  \bibinfo{pages}{1--52}.
\newblock
\showISBNx{978-1-4503-8096-6}
\urldef\tempurl%
\url{https://doi.org/10.1145/3411764.3445315}
\showDOI{\tempurl}


\bibitem[Price~II(2018)]%
        {priceii2018medical}
\bibfield{author}{\bibinfo{person}{W.~Nicholson Price~II}.}
  \bibinfo{year}{2018}\natexlab{}.
\newblock \showarticletitle{Medical {{Malpractice}} and {{Black-box
  Medicine}}}.
\newblock In \bibinfo{booktitle}{\emph{Big {{Data}}, {{Health Law}}, and
  {{Bioethics}}}}. \bibinfo{publisher}{{Cambridge University PRess}}.
\newblock


\bibitem[Raji et~al\mbox{.}(2020)]%
        {raji2020closing}
\bibfield{author}{\bibinfo{person}{Inioluwa~Deborah Raji},
  \bibinfo{person}{Andrew Smart}, \bibinfo{person}{Rebecca~N. White},
  \bibinfo{person}{Margaret Mitchell}, \bibinfo{person}{Timnit Gebru},
  \bibinfo{person}{Ben Hutchinson}, \bibinfo{person}{Jamila {Smith-Loud}},
  \bibinfo{person}{Daniel Theron}, {and} \bibinfo{person}{Parker Barnes}.}
  \bibinfo{year}{2020}\natexlab{}.
\newblock \showarticletitle{Closing the {{AI}} Accountability Gap: Defining an
  End-to-End Framework for Internal Algorithmic Auditing}. In
  \bibinfo{booktitle}{\emph{Proceedings of the 2020 {{Conference}} on
  {{Fairness}}, {{Accountability}}, and {{Transparency}}}}.
  \bibinfo{publisher}{{ACM}}, \bibinfo{address}{{Barcelona Spain}},
  \bibinfo{pages}{33--44}.
\newblock
\showISBNx{978-1-4503-6936-7}
\urldef\tempurl%
\url{https://doi.org/10.1145/3351095.3372873}
\showDOI{\tempurl}


\bibitem[Raney(2019)]%
        {raney2019guest}
\bibfield{author}{\bibinfo{person}{Gary Raney}.}
  \bibinfo{year}{2019}\natexlab{}.
\newblock \showarticletitle{Guest Opinion: {{Using}} Risk Assessments in the
  Criminal Justice System}.
\newblock \bibinfo{journal}{\emph{The Idaho Press}} (\bibinfo{date}{Feb.}
  \bibinfo{year}{2019}).
\newblock
\urldef\tempurl%
\url{https://www.idahopress.com/opinion/guest_opinions/guest-opinion-using-risk-assessments-in-the-criminal-justice-system/article_acd4f7ec-9dde-54b6-a3c3-d4e35451b5de.html}
\showURL{%
\tempurl}


\bibitem[Rao et~al\mbox{.}(2016)]%
        {rao2016transition}
\bibfield{author}{\bibinfo{person}{Shebani Rao}, \bibinfo{person}{Kevin
  Warwick}, \bibinfo{person}{Gary Christensen}, {and} \bibinfo{person}{Colleen
  Owens}.} \bibinfo{year}{2016}\natexlab{}.
\newblock \bibinfo{booktitle}{\emph{Transition from {{Jail}} to {{Community}}
  ({{TJC}}) {{Initiative}}}}.
\newblock \bibinfo{type}{{T}echnical {R}eport}. \bibinfo{institution}{{Urban
  Institute}}.
\newblock


\bibitem[Rice and Swesnik(2013)]%
        {rice2013discriminatory}
\bibfield{author}{\bibinfo{person}{Lisa Rice} {and} \bibinfo{person}{Deidre
  Swesnik}.} \bibinfo{year}{2013}\natexlab{}.
\newblock \showarticletitle{Discriminatory {{Effects}} of {{Credit Scoring}} on
  {{Communities}} of {{Color}}}.
\newblock \bibinfo{journal}{\emph{Suffolk University Law Review}}
  \bibinfo{volume}{46} (\bibinfo{year}{2013}).
\newblock


\bibitem[Rudin(2019)]%
        {rudin2019stop}
\bibfield{author}{\bibinfo{person}{Cynthia Rudin}.}
  \bibinfo{year}{2019}\natexlab{}.
\newblock \showarticletitle{Stop Explaining Black Box Machine Learning Models
  for High Stakes Decisions and Use Interpretable Models Instead}.
\newblock \bibinfo{journal}{\emph{Nature Machine Intelligence}}
  \bibinfo{volume}{1}, \bibinfo{number}{5} (\bibinfo{date}{May}
  \bibinfo{year}{2019}), \bibinfo{pages}{206--215}.
\newblock
\showISSN{2522-5839}
\urldef\tempurl%
\url{https://doi.org/10.1038/s42256-019-0048-x}
\showDOI{\tempurl}


\bibitem[Sheehan and Du(2022)]%
        {sheehan2022what}
\bibfield{author}{\bibinfo{person}{Matt Sheehan} {and} \bibinfo{person}{Sharon
  Du}.} \bibinfo{year}{2022}\natexlab{}.
\newblock \bibinfo{title}{What {{China}}'s {{Algorithm Registry Reveals}} about
  {{AI Governance}}}.
\newblock
\newblock
\urldef\tempurl%
\url{https://carnegieendowment.org/2022/12/09/what-china-s-algorithm-registry-reveals-about-ai-governance-pub-88606}
\showURL{%
\tempurl}


\bibitem[Sherif(2016)]%
        {sherif2016beginning}
\bibfield{author}{\bibinfo{person}{Nazneen Sherif}.}
  \bibinfo{year}{2016}\natexlab{}.
\newblock \showarticletitle{The Beginning of the End for Footloose Modelling}.
\newblock \bibinfo{journal}{\emph{Risk.net}} (\bibinfo{date}{June}
  \bibinfo{year}{2016}).
\newblock
\urldef\tempurl%
\url{https://www.risk.net/risk-management/2463006/beginning-end-footloose-modelling}
\showURL{%
\tempurl}


\bibitem[Siegmann and Anderljung(RINT)]%
        {siegmann2022brussels}
\bibfield{author}{\bibinfo{person}{Charlotte Siegmann} {and}
  \bibinfo{person}{Markus Anderljung}.} \bibinfo{year}{July 2022 [AUTHOR
  PREPRINT]}\natexlab{}.
\newblock \bibinfo{booktitle}{\emph{The {{Brussels Effect}} and {{Artificial
  Intelligence}}: {{Will EU}} Regulation Shape the Global Artificial
  Intelligence Market?}}
\newblock \bibinfo{type}{{T}echnical {R}eport}. \bibinfo{institution}{{Centre
  for the Governance of AI}}.
\newblock


\bibitem[{State of Colorado}(2022)]%
        {stateofcolorado2022division}
\bibfield{author}{\bibinfo{person}{{State of Colorado}}.}
  \bibinfo{year}{2022}\natexlab{}.
\newblock \bibinfo{title}{Division of {{Insurance Partners}} with {{ORCAA}} to
  {{Protect Colorado Insurance Consumers}}}.
\newblock
\newblock
\urldef\tempurl%
\url{https://doi.colorado.gov/news-releases-consumer-advisories/division-of-insurance-partners-with-orcaa-to-protect-colorado}
\showURL{%
\tempurl}


\bibitem[States(1974)]%
        {unitedstates1974equal}
\bibfield{author}{\bibinfo{person}{United States}.}
  \bibinfo{year}{1974}\natexlab{}.
\newblock \bibinfo{title}{Equal {{Credit Opportunity Act}}}.
\newblock
\newblock
\urldef\tempurl%
\url{https://www.govinfo.gov/content/pkg/USCODE-2011-title15/html/USCODE-2011-title15-chap41-subchapIV.htm}
\showURL{%
\tempurl}


\bibitem[Tomsett et~al\mbox{.}(2018)]%
        {tomsett2018interpretable}
\bibfield{author}{\bibinfo{person}{Richard Tomsett}, \bibinfo{person}{Dave
  Braines}, \bibinfo{person}{Dan Harborne}, \bibinfo{person}{Alun Preece},
  {and} \bibinfo{person}{Supriyo Chakraborty}.}
  \bibinfo{year}{2018}\natexlab{}.
\newblock \showarticletitle{Interpretable to {{Whom}}? {{A Role-based Model}}
  for {{Analyzing Interpretable Machine Learning Systems}}}. In
  \bibinfo{booktitle}{\emph{Proc. 2018 {{ICML Workshop}} on {{Human
  Interpretability}} in {{Machine Learning}}}}. \bibinfo{address}{{Stockholm,
  Sweden}}.
\newblock


\bibitem[Tournas and Bowman(2021)]%
        {tournas2021ai}
\bibfield{author}{\bibinfo{person}{Lucille~Nalbach Tournas} {and}
  \bibinfo{person}{Diana~M. Bowman}.} \bibinfo{year}{2021}\natexlab{}.
\newblock \showarticletitle{{{AI Insurance}}: {{Risk Management}} 2.0}.
\newblock \bibinfo{journal}{\emph{IEEE Technology and Society Magazine}}
  \bibinfo{volume}{40}, \bibinfo{number}{4} (\bibinfo{date}{Dec.}
  \bibinfo{year}{2021}), \bibinfo{pages}{52--56}.
\newblock
\showISSN{0278-0097, 1937-416X}
\urldef\tempurl%
\url{https://doi.org/10.1109/MTS.2021.3123750}
\showDOI{\tempurl}


\bibitem[UNESCO(2021)]%
        {unesco2021recommendation}
\bibfield{author}{\bibinfo{person}{UNESCO}.} \bibinfo{year}{2021}\natexlab{}.
\newblock \bibinfo{booktitle}{\emph{Recommendation on the {{Ethics}} of
  {{Artificial Intelligence}}}}.
\newblock \bibinfo{type}{{T}echnical {R}eport} SHS/BIO/PI/2021/1.
  \bibinfo{institution}{{UNESCO}}.
\newblock
\urldef\tempurl%
\url{https://unesdoc.unesco.org/ark:/48223/pf0000381137}
\showURL{%
\tempurl}


\bibitem[Upbin(2020)]%
        {upbin2020how}
\bibfield{author}{\bibinfo{person}{Bruce Upbin}.}
  \bibinfo{year}{2020}\natexlab{}.
\newblock \bibinfo{title}{How {{Machine Learning Is}} ({{And Isn}}'t)
  {{Changing Fair Lending}}}.
\newblock
\newblock
\urldef\tempurl%
\url{https://www.zest.ai/insights/how-machine-learning-is-and-isnt-changing-fair-lending}
\showURL{%
\tempurl}


\bibitem[{U.S. National Institute of Standards {and} Technology}(2023)]%
        {u.s.nationalinstituteofstandardsandtechnology2023artificial}
\bibfield{author}{\bibinfo{person}{{U.S. National Institute of Standards {and}
  Technology}}.} \bibinfo{year}{2023}\natexlab{}.
\newblock \bibinfo{title}{Artificial {{Intelligence Risk Management Framework}}
  ({{AI RMF}} 1.0)}.
\newblock
\newblock
\urldef\tempurl%
\url{https://doi.org/10.6028/NIST.AI.100-1}
\showURL{%
\tempurl}


\bibitem[{U.S. Office of the Comptroller of the Currency}(2021)]%
        {u.s.officeofthecomptrollerofthecurrency2021comptroller}
\bibfield{author}{\bibinfo{person}{{U.S. Office of the Comptroller of the
  Currency}}.} \bibinfo{year}{2021}\natexlab{}.
\newblock \bibinfo{title}{Comptroller's {{Handbook}}: {{Model Risk
  Management}}}.
\newblock
\newblock
\urldef\tempurl%
\url{https://www.occ.treas.gov/publications-and-resources/publications/comptrollers-handbook/files/model-risk-management/pub-ch-model-risk.pdf}
\showURL{%
\tempurl}


\bibitem[Veale and Zuiderveen~Borgesius(2021)]%
        {veale2021demystifying}
\bibfield{author}{\bibinfo{person}{Michael Veale} {and}
  \bibinfo{person}{Frederik Zuiderveen~Borgesius}.}
  \bibinfo{year}{2021}\natexlab{}.
\newblock \showarticletitle{Demystifying the {{Draft EU Artificial Intelligence
  Act}} \textemdash{} {{Analysing}} the Good, the Bad, and the Unclear Elements
  of the Proposed Approach}.
\newblock \bibinfo{journal}{\emph{Computer Law Review International}}
  \bibinfo{volume}{22}, \bibinfo{number}{4} (\bibinfo{date}{Aug.}
  \bibinfo{year}{2021}), \bibinfo{pages}{97--112}.
\newblock
\showISSN{2194-4164}
\urldef\tempurl%
\url{https://doi.org/10.9785/cri-2021-220402}
\showDOI{\tempurl}


\bibitem[Washington(2018)]%
        {washington2018how}
\bibfield{author}{\bibinfo{person}{Anne~L. Washington}.}
  \bibinfo{year}{2018}\natexlab{}.
\newblock \showarticletitle{How to {{Argue}} with an {{Algorithm}}: {{Lessons}}
  from the {{COMPAS-ProPublica Debate}}}.
\newblock \bibinfo{journal}{\emph{Colorado Technology Law Journal}}
  \bibinfo{volume}{17}, \bibinfo{number}{1} (\bibinfo{year}{2018}).
\newblock


\bibitem[Wendehorst(2022)]%
        {wendehorst2022ai}
\bibfield{author}{\bibinfo{person}{Christiane Wendehorst}.}
  \bibinfo{year}{2022}\natexlab{}.
\newblock \bibinfo{booktitle}{\emph{{{AI}} Liability in {{Europe}}:
  {{Anticipating}} the {{EU AI Liability Directive}}}}.
\newblock \bibinfo{type}{{T}echnical {R}eport}. \bibinfo{institution}{{Ada
  Lovelace Institute}}.
\newblock
\urldef\tempurl%
\url{https://www.adalovelaceinstitute.org/wp-content/uploads/2022/09/Ada-Lovelace-Institute-Expert-Explainer-AI-liability-in-Europe.pdf}
\showURL{%
\tempurl}


\bibitem[Wex(2022)]%
        {wex2022ultrahazardous}
\bibfield{author}{\bibinfo{person}{Wex}.} \bibinfo{year}{2022}\natexlab{}.
\newblock \bibinfo{title}{Ultrahazardous Activity}.
\newblock
\newblock
\urldef\tempurl%
\url{https://www.law.cornell.edu/wex/proximate_cause}
\showURL{%
\tempurl}


\bibitem[Whittaker et~al\mbox{.}(2018)]%
        {whittaker2018ai}
\bibfield{author}{\bibinfo{person}{Meredith Whittaker}, \bibinfo{person}{Kate
  Crawford}, \bibinfo{person}{Roel Dobbe}, \bibinfo{person}{Genevieve Fried},
  \bibinfo{person}{Elizabeth Kaziunas}, \bibinfo{person}{Varoon Mathur},
  \bibinfo{person}{Sarah Myers~West}, \bibinfo{person}{Rashida Richardson},
  \bibinfo{person}{Jason Schultz}, {and} \bibinfo{person}{Oscar Schwartz}.}
  \bibinfo{year}{2018}\natexlab{}.
\newblock \bibinfo{booktitle}{\emph{{{AI Now Report}} 2018}}.
\newblock \bibinfo{type}{{T}echnical {R}eport}. \bibinfo{institution}{{New York
  University AI Now Institute}}.
\newblock
\urldef\tempurl%
\url{https://ainowinstitute.org/AI_Now_2018_Report.pdf}
\showURL{%
\tempurl}


\bibitem[Williams(2008)]%
        {williams2008fair}
\bibfield{author}{\bibinfo{person}{Orice~M. Williams}.}
  \bibinfo{year}{2008}\natexlab{}.
\newblock \bibinfo{booktitle}{\emph{Fair {{Lending}}: {{Race}} and {{Gender
  Data}} Are {{Limited}} for {{Nonmortgage Lending}}}}.
\newblock \bibinfo{type}{{T}echnical {R}eport} GAO-08-698.
  \bibinfo{institution}{{U.S. Government Accountability Office}}.
\newblock
\urldef\tempurl%
\url{https://www.gao.gov/products/gao-08-698}
\showURL{%
\tempurl}


\bibitem[Zerilli et~al\mbox{.}(2022)]%
        {zerilli2022how}
\bibfield{author}{\bibinfo{person}{John Zerilli}, \bibinfo{person}{Umang
  Bhatt}, {and} \bibinfo{person}{Adrian Weller}.}
  \bibinfo{year}{2022}\natexlab{}.
\newblock \showarticletitle{How Transparency Modulates Trust in Artificial
  Intelligence}.
\newblock \bibinfo{journal}{\emph{Patterns}} \bibinfo{volume}{3},
  \bibinfo{number}{4} (\bibinfo{date}{April} \bibinfo{year}{2022}),
  \bibinfo{pages}{100455}.
\newblock
\showISSN{26663899}
\urldef\tempurl%
\url{https://doi.org/10.1016/j.patter.2022.100455}
\showDOI{\tempurl}


\bibitem[Zhang et~al\mbox{.}(2022)]%
        {zhang2022explainable}
\bibfield{author}{\bibinfo{person}{Chanyuan~(Abigail) Zhang},
  \bibinfo{person}{Soohyun Cho}, {and} \bibinfo{person}{Miklos Vasarhelyi}.}
  \bibinfo{year}{2022}\natexlab{}.
\newblock \showarticletitle{Explainable {{Artificial Intelligence}} ({{XAI}})
  in Auditing}.
\newblock \bibinfo{journal}{\emph{International Journal of Accounting
  Information Systems}}  \bibinfo{volume}{46} (\bibinfo{date}{Sept.}
  \bibinfo{year}{2022}), \bibinfo{pages}{100572}.
\newblock
\showISSN{14670895}
\urldef\tempurl%
\url{https://doi.org/10.1016/j.accinf.2022.100572}
\showDOI{\tempurl}


\end{thebibliography}

\end{document}